\documentclass[10pt, english, oneside, a4paper]{article}

\usepackage[utf8]{inputenc}
\usepackage{bm}
\usepackage[style=american]{csquotes}
\usepackage{rotating}
\usepackage[hidelinks]{hyperref}
\usepackage[left=1.25in,right=1.25in,top=1.25in,bottom=1.25in]{geometry} 
\usepackage{etoolbox}
\usepackage{tcolorbox}
\usepackage{enumitem}
\usepackage{scrextend}
\usepackage{centernot}
\usepackage{lipsum}
\usepackage{amsfonts}
\usepackage{amssymb}
\usepackage{amsmath}
\usepackage{amsthm}
\usepackage{amsopn}
\usepackage{graphicx}
\usepackage{epstopdf}
\usepackage{algorithmic}
\usepackage{benstyle_arXiv}
\usepackage{tikz}
\usepackage{bbm}
\usepackage{mathtools}
\usepackage{array}
\usepackage{rotating}
\usepackage{booktabs}
\usepackage[normalem]{ulem}

\usepackage{caption}
\usetikzlibrary{calc}
\usetikzlibrary{decorations.pathreplacing,calc}

\ifpdf
  \DeclareGraphicsExtensions{.eps,.pdf,.png,.jpg}
\else
  \DeclareGraphicsExtensions{.eps}
\fi

\numberwithin{equation}{section}

\theoremstyle{plain}
\newtheorem{theorem}{Theorem}[section]

\newtheorem{mainresult}[theorem]{Main result}

\theoremstyle{definition}
\newtheorem{definition}[theorem]{Definition}

\theoremstyle{remark}
\newtheorem{remark}[theorem]{Remark}

\def\C{\mathbb{C}}

\def\R{\mathbb{R}}

\renewcommand\phi{\varphi}

\newcommand{\eps}{\epsilon}

\renewcommand{\ge}{\geqslant}
\renewcommand{\le}{\leqslant}
\renewcommand{\geq}{\geqslant}
\renewcommand{\leq}{\leqslant}
\renewcommand{\hat}{\widehat}
\renewcommand{\tilde}{\widetilde}

\newcommand{\vertiii}[1]{{\left\vert\kern-0.25ex\left\vert\kern-0.25ex\left\vert #1 
    \right\vert\kern-0.25ex\right\vert\kern-0.25ex\right\vert}}

\usepackage{enumitem}
\usepackage[percent]{overpic}
\usepackage{pict2e}

\usepackage{amsopn}
\usepackage{cite}

\AtBeginEnvironment{figure}{\setlength{\tabcolsep}{2pt}}

\usepackage{rotating}
\tikzstyle{stuff_fill}=[rectangle,draw,fill=pink,minimum size=0.5em]
\usetikzlibrary{calc}
\usetikzlibrary{decorations.pathreplacing,calc}
\newcounter{arrow}
\title{The troublesome kernel \\---\\ On hallucinations, no free lunches and the accuracy-stability trade-off in inverse problems\thanks{
NMG acknowledges support from a UK EPSRC grant. ACH acknowledges support from a Royal Society University
Research Fellowship and the Leverhulme Prize 2017.
BA acknowledges the support of the PIMS CRG ``High-dimensional Data
Analysis'', SFU's Big Data Initiative ``Next Big Question" Fund and NSERC
through grant R611675. }}

\author{
    Nina M. Gottschling\thanks{University of Cambridge, Wilberforce Road, Cambridge CB3 0WA, UK (\href{mailto:nmg43@cam.ac.uk}{nmg43@cam.ac.uk})}
\and Vegard Antun\thanks{University of Oslo, P.O box 1053, Blindern, 0316 Oslo,   Norway (\href{mailto:vegarant@math.uio.no}{vegarant@math.uio.no})}
\and Anders C. Hansen\thanks{University of Cambridge, Wilberforce Road, Cambridge CB3 0WA,  UK  (\href{mailto:ach70@cam.ac.uk}{ach70@cam.ac.uk})}
\and Ben Adcock\thanks{Simon Fraser University, 8888 University Drive
Burnaby, BC V5A 1S6, Canada (\href{mailto:ben\_adcock@sfu.ca}{ben\_adcock@sfu.ca})}
}

\usepackage{amsopn}

\ifpdf
\hypersetup{
  pdftitle={The troublesome kernel},
  pdfauthor={Nina M. Gottschling, Vegard Antun, Anders C. Hansen and Ben Adcock}
}
\fi

\makeatletter
\renewenvironment*{displayquote}
  {\begingroup\setlength{\leftmargini}{0.6cm}\csq@getcargs{\csq@bdquote{}{}}}
  {\csq@edquote\endgroup}
\makeatother

\begin{document}

\maketitle

\begin{abstract}
Methods inspired by Artificial Intelligence (AI) are starting to fundamentally change computational science and engineering through breakthrough performances on challenging problems. However, reliability and trustworthiness of such techniques is a major concern. In inverse problems in imaging, the focus of this paper, there is increasing empirical evidence that methods may suffer from hallucinations, i.e., false, but realistic-looking artifacts; instability, i.e., sensitivity to perturbations in the data; and unpredictable generalization, i.e., excellent performance on some images, but significant deterioration on others. This paper provides a theoretical foundation for these phenomena. We give mathematical explanations for how and when such effects arise in arbitrary reconstruction methods, with several of our results taking the form of `no free lunch' theorems. Specifically, we show that (i) methods that overperform on a single image can wrongly transfer details from one image to another, creating a hallucination, (ii) methods that overperform on two or more images can hallucinate or be unstable, (iii) optimizing the accuracy-stability trade-off is generally difficult, (iv) hallucinations and instabilities, if they occur, are not rare events, and may be encouraged by standard training, (v) it may be impossible to construct optimal reconstruction maps for certain problems. Our results trace these effects to the kernel of the forward operator whenever it is nontrivial, but also apply to the case when the forward operator is ill-conditioned. Based on these insights, our work aims to spur  research into new ways to develop robust and reliable AI-based methods for inverse problems in imaging.
\end{abstract}

\paragraph*{Keywords}
Inverse problems, imaging, deep learning, hallucinations, instability, no-free lunch theorems

\paragraph*{Mathematics Subject Classification (2010):}
65R32, 94A08, 68T05, 65M12

\section{Introduction}\label{sec:introduction}

It is impossible to overstate the impact that Neural Networks (NNs) and Deep Learning (DL) have had in recent years in Machine Learning (ML) applications such as image classification, speech recognition and natural language processing.  Perhaps unsurprisingly, the development and use of Artificial Intelligence (AI)-inspired methods for challenging problems in the computational sciences has recently become an active area of inquiry.  Areas of particular notice include numerical PDEs \cite{raissi2019physics-informed}, discovering PDE dynamics \cite{rudy2017data-driven}, Uncertainty Quantification \cite{geist2021numerical,bhattacharya2021model} and high-dimensional approximation \cite{adcock2021deep,adcock2021gap}. 

Arguably, however, the area of computational science in which AI-based methods have been most actively investigated is inverse problems in imaging. The task of recovering an image from measurements  is of vital importance in a wide range of scientific, industrial and medical applications. These include, but are by no means limited to, electron and fluorescence microscopy, seismic imaging, Nuclear Magnetic Resonance (NMR), Magnetic Resonance Imaging (MRI) and X-Ray Computerized Tomography (CT).  In the last several years, there has been an unprecedented amount of activity in the application of ML, and specifically, DL, in this area (see \S \ref{s:relwork} for an overview of relevant literature). Given the potential for breakthrough performance, it seems possible that the future of the field lies with AI-inspired algorithms. Notably, their potential has been described by \textit{Nature} as `transformative' \cite{Str-18}.\footnote{To be specific, \cite{Str-18} is titled `AI transforms image reconstruction' and features a new DL approach \cite{Bo-18} which `improves speed, accuracy and robustness of biomedical image reconstruction'.}

\subsection{Hallucinations, instability and unpredictable performance}\label{ss:growing-concerns}
\begin{figure}[p]
	\begin{center}
		\setlength{\tabcolsep}{1pt} 
		\begin{footnotesize}
			\begin{tabular}{@{}>{\centering}m{0.14\textwidth}>{\centering}m{0.27\textwidth}>{\centering}m{0.27\textwidth}>{\centering\arraybackslash}m{0.27\textwidth}@{}}   
				&Original  $|x|$ & Original $|x|$ (crop) & NN $\Psi(Ax)$ (crop) \\
				\begin{turn}{90}\parbox[b]{0.28\textwidth}{\begin{center}\begin{footnotesize}\textbf{Parallel MRI \\ Noiseless} \\ $320\times 320$, $\quad$ 8$\times$accel. \\ Variational network (baseline fastMRI20 \cite{fastmri20})\end{footnotesize}\end{center}}\end{turn}
				&\includegraphics[width=\linewidth]{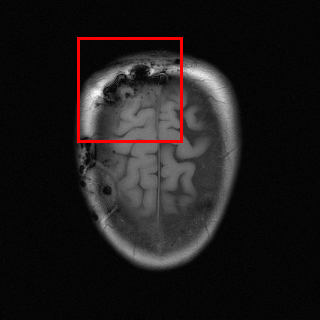}
				&\includegraphics[width=\linewidth]{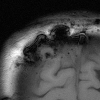}
				&\includegraphics[width=\linewidth]{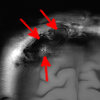}\\
				&Original  $|x|$ (crop) & NN $\Psi(Ax)$ (crop) & Res.\ $|\Psi(Ax)-x|$ (crop) \\
				\begin{turn}{90}\parbox[b]{0.28\textwidth}{\begin{center}\begin{footnotesize}\textbf{Parallel MRI \\ Noiseless} \\ $320\times 320$, $\quad$ 4$\times$accel. \\ XPDNet \cite{ramzi2021xpdnet}  \\ (Figure from \cite{fastmri20})\end{footnotesize}\end{center}}\end{turn}
				&\includegraphics[width=\linewidth]{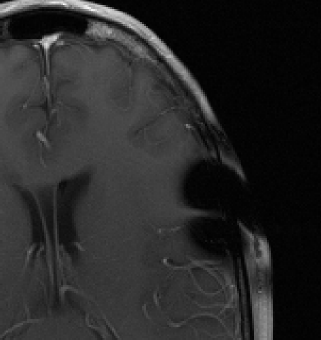}
				&\includegraphics[width=\linewidth]{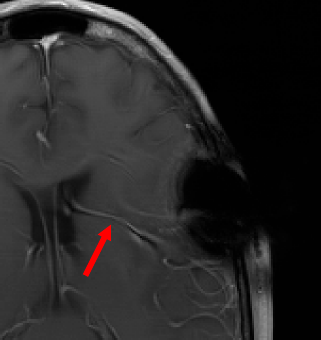}
				&\includegraphics[width=\linewidth]{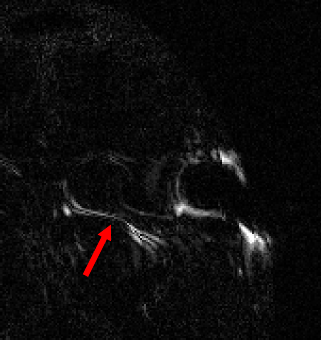}\\
				&Original  $x$ & Original $x$ (crop) & NN $\Psi(Ax)$ (crop) \\
				\begin{turn}{90}
					\parbox[b]{0.28\textwidth}{\begin{center}\begin{footnotesize}\textbf{Optical microscopy\\ Noiseless} \\ $512\times 512$, $\quad$ 2$\times$diff. limit. \\ DFGAN-SISR \cite{qiao2021evaluation} \end{footnotesize}\end{center}}
				\end{turn}
				&\includegraphics[width=\linewidth]{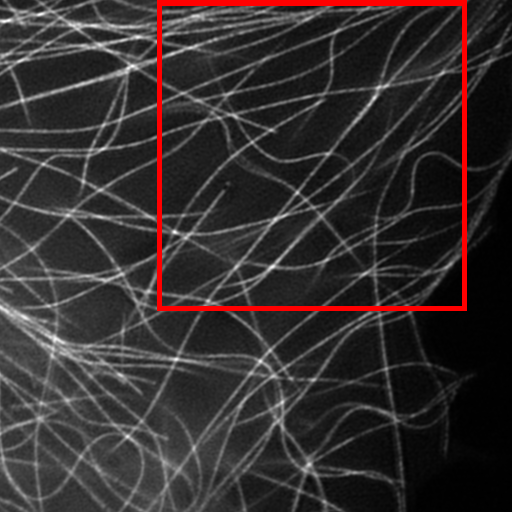}
				&\includegraphics[width=\linewidth]{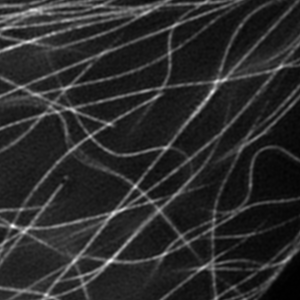}
				&\includegraphics[width=\linewidth]{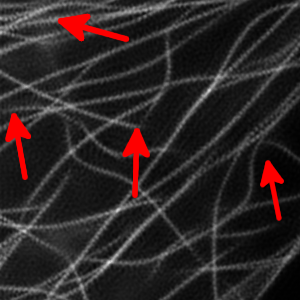} \\
				&Original  $x$ & Original $x$ (crop) & NN $\Psi(Ax)$ (crop) \\
				\begin{turn}{90}
					\parbox[b]{0.28\textwidth}{\begin{center}\begin{footnotesize}\textbf{Fluorescence  micros. \\ Noiseless} \\ $512\times 512$, $\quad$ 20$\times$ accel \\ Learned inv.\ + Tiramisunet  \end{footnotesize}\end{center}}
				\end{turn}
				&\includegraphics[width=\linewidth]{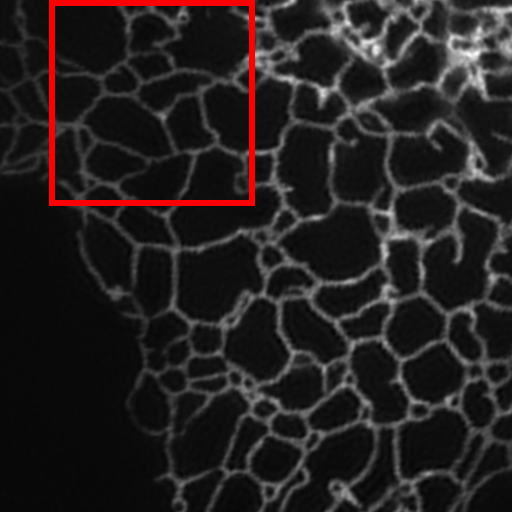}
				&\includegraphics[width=\linewidth]{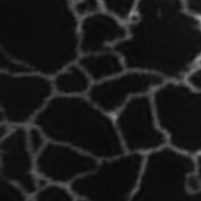}
				&
			\begin{overpic}[width=\linewidth]{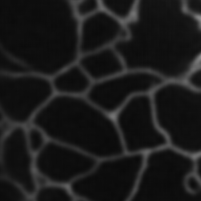}
				\linethickness{1.4pt}
				\put(46,96){\color{red}\vector(1.2,-1){12}}
				\put(35,79){\color{red}\vector(1,0){17}}
				\put(88,92){\color{red}\vector(-1,0){17}}
			\end{overpic} \\
			\end{tabular}
		\end{footnotesize}
	\end{center}
    \caption{\label{fig:1} (\textbf{AI-generated hallucinations in different imaging modalities}). Trained NNs $\Psi \colon \C^{m} \to \R^{N}$ for different imaging modalities generate hallucinations, i.e., realistic-looking artifacts, when evaluated on test data. Note that $m$, $N$ and $A$ vary between the experiments. In the first three rows we consider trained NNs from the cited publications. In row four, we have trained a NN using data from \cite{qiao2021evaluation}. For more information on the training procedure see \S \ref{s:figures-experiments}.} 
\end{figure}

However, there is increasing awareness that methods in inverse problems  can suffer from (i) \textit{hallucinations}, i.e., realistic-looking artifacts that appear in a reconstructed image that are not present in the ground truth image; (ii) \textit{instabilities}, i.e., sensitivity to perturbations in the measurements; and (iii) \textit{unpredictable performance}, i.e., excellent performance on some images, but significantly worse performance on other nearby images. For example, from the evaluation of the \textit{2020 Facebook fastMRI challenge} \cite{fastmri20}:
\begin{displayquote}
    \enquote{\textit{Such hallucinatory features are not acceptable and especially problematic if they mimic normal structures that are either not present or actually abnormal. Neural network models can be unstable as demonstrated via adversarial perturbation studies \cite{Anetal-19}.}}
\end{displayquote}
\noindent Similarly, in the work \textit{On hallucinations in tomographic image reconstruction} \cite{Bha20}:
\begin{displayquote}
    \enquote{\textit{The potential lack of generalization of deep learning-based
reconstruction methods as well as their innate unstable nature
may cause false structures to appear in the reconstructed
image that are absent in the object being imaged.}}
\end{displayquote}
\noindent Nor are these issues limited to medical imaging. In \textit{Applications, promises, and pitfalls of deep learning for fluorescence image reconstruction} \cite{bel19}, the authors write
\begin{displayquote}
    \enquote{\textit{The most serious issue when applying deep learning for discovery is that of hallucination. [...] These hallucinations are deceptive artifacts that appear highly plausible in the absence of contradictory information and can be challenging, if not impossible, to detect.}} 
\end{displayquote}
\noindent and in \textit{The promise and peril of deep learning in microscopy} \cite{hoff21}, they write
\begin{displayquote}
    \enquote{\textit{However, if the neural network encounters unknown specimens, or known specimens imaged with unknown microscopes, it can produce nonsensical results.}}
\end{displayquote}
\noindent Similar commentary can also be found in \cite{laine2021avoiding,  liu2022medical, morshuis2022adversarial, lustig_pnas22, mean_field_pmlr21, varoquaux2022machine, yu2022validation, wu2021medical} and references therein. 
To highlight this issue, in Fig.\ \ref{fig:1} we present several examples of AI-based methods for different imaging tasks. In all cases, the recovered images exhibit realistic-looking features that are not present in the corresponding ground truth images; that is to say, they all generate hallucinations.

\subsection{This paper: hallucinations, no free lunches and the accuracy-stability trade-off}
Because of these concerns, there is now a growing research focus on empirically examining the robustness (or lack thereof) and performance of AI-based methods in practical imaging scenarios \cite{huang18,Anetal-19,johnson2021evaluation,fastmri20,morshuis2022adversarial,lustig_pnas22,yu2022validation,Heckel21,genzel2020solving,zhang2021instabilities}. 
The aim of this paper is to provide a theoretical counterpart to these studies. We do so by developing a mathematical understanding for how and why such phenomena arise. Our main results trace the source of these phenomena to the kernel of the forward (sampling) operator, which is often nontrivial (and large) in practical imaging scenarios. However, our results also apply to problems with forward operators that have trivial kernels but are ill-conditioned. As we discuss in more detail in \S \ref{s:overview}, our results describe mechanisms that can cause hallucinations, instabilities or unpredictable performance to occur for such forward operators. We also present a series of numerical experiments that illustrate these mechanisms in practice.

Our results follow the grand tradition of `no free lunch' theorems in ML  \cite{ML_book}. Broadly speaking, our results show that a reconstruction procedure that \textit{overperforms} in a certain sense, must inevitably succumb to one or more of the main phenomena. Several of these results also describe a fundamental \textit{accuracy-stability trade-off} for inverse problems, wherein if the accuracy of a method is pushed too far (e.g., by driving the training error to zero), it inevitably becomes unstable.

\begin{remark}[Are these phenomena exclusive to AI-based methods?]
\label{rem:sparse-reg-also}
The increasing focus on the pitfalls of DL for imaging has also led researchers in recent years to re-examine standard methods -- for example, those based on sparse regularization \cite{adcock2021compressive} -- more closely from the perspective of performance versus undesirable effects such as hallucinations, instabilities and unpredictable performance. Several works have reported that these methods may also be susceptible to some of these issues \cite{genzel2020solving,zhang2021instabilities, Heckel21,alaifari2022localized}, albeit arguably in not as dramatic ways that certain AI-based methods can be. We discuss this matter in more detail in \S \ref{s:relwork}.

The purpose of this work is not to advocate one methodology over the other. Rather, our aim is to develop a series of the theoretical mechanisms that inevitably lead to such undesirable effects. While some of our results are geared towards learning-based methods, most of them apply to arbitrary reconstruction methods, thereby including sparse regularization as a particular case. For example, the accuracy-stability trade-off that we establish -- an effect that was observed empirically in several of the aforementioned works \cite{zhang2021instabilities,genzel2020solving}  -- states that \textit{any} method, AI-based or not, that strives to achieve too high accuracy must inevitably become unstable.
\end{remark}

\subsection{Robustness and trustworthiness in AI}

Our work is part of the broader discussion on robustness (or lack thereof) and trustworthiness of AI. This is not just an issue in the computational sciences, but one which affects all sectors in which AI-based techniques are beginning to be actively used. It is notable that the governmental bodies are currently striving to address these concerns. For instance, the European Commission is in the process of outlining a legal framework for the use of AI, with an emphasis on robustness and trustworthiness \cite{EU_AI_regulations}. 

Thus, the performance of DL in inverse problems in imaging is a part of a much larger issue that needs addressing at all levels. The aim of this paper is not to provide solutions to these issues. Yet by exposing the underlying mechanisms that cause undesirable behaviour, we aim to gain insight into how these issues may be eventually overcome, thus enabling the safe and trustworthy use of DL in computational sciences.

\section{Overview of the paper}\label{s:overview}

In this section, we give an overview of the paper. We first formalize the main problem studied, then we give a summary of our main results. Finally, we conclude with a discussion of related literature.

\subsection{Problem outline}

The concern of this paper is the discrete inverse problem
\begin{equation}
	\label{eq:invintro}
	\mbox{given measurements $y = A x+e$, recover $x$.}
\end{equation}
Here $A \in \bbC^{m \times N}$ is the \textit{sampling operator} (also called \textit{measurement matrix}), $y \in \bbC^m$ is a vector of \textit{measurements}, $e \in \bbC^m$ is measurement noise  and $x \in \bbC^N$ is the (unknown) object to recover (typically a discrete image in a vectorized form). While seemingly simple, the model \eqref{eq:invintro} is often sufficient to model many applications, including all of those mentioned above. 
In many imaging scenarios (e.g., undersampled MRI),  $A$ may satisfy $\mathrm{rank}(A) < N$, which makes its kernel $\cN(A)$ nontrivial. Note that this is always the case when $m < N$, i.e., when there are fewer measurements $m$ than unknowns $N$. This makes solving \eqref{eq:invintro} a challenge, since,  even in the noiseless case $e = 0$, there are infinitely many candidate solutions $x$ that yield the same measurements $y$. In other settings (e.g., image deblurring), $A$ may have $\mathrm{rank}(A) = N$, i.e., $\cN(A) = \{ 0 \}$, but be ill-conditioned, thus making sensitivity to noise an issue for any recovery procedure.

This paper is about \textit{reconstruction maps} for \eqref{eq:invintro}. These are mappings of the form $\Psi \colon \bbC^m \rightarrow \bbC^N$ from the measurement domain $\bbC^m$ to the object domain $\bbC^N$. Occasionally, we also allow for multivalued maps, which we denote as $\Psi \colon \bbC^m \rightrightarrows \bbC^N$.
To design a reconstruction map for \eqref{eq:invintro}, one normally assumes that the desired images $x$ belong to some set $\mathcal{M}_1 \subset\C^N$. This set is sometimes referred to as a \textit{model class} or \textit{image manifold}.
Thus, rather than solving \eqref{eq:invintro}, one solves the problem
\begin{equation}
	\label{eq:invintro2}
    \mbox{Given measurements $y = A x+e$ of $x \in \mathcal{M}_1$, recover $x$.}
\end{equation}
Broadly speaking, methods for solving \eqref{eq:invintro2} can be divided into two types:

\begin{enumerate}[label=(\roman*),leftmargin=0.75cm]
    \item In \textit{model-based} methods, one makes explicit assumptions about $\mathcal{M}_1$, and designs the reconstruction map based on the choice of $\mathcal{M}_1$. Common choices for $\mathcal{M}_1$ include sets of images which are approximately sparse in a wavelet basis (or some other multiscale system such as curvelets or shearlets) or images with approximately sparse gradient. To recover $x \in \cM_1$, one then usually solves a regularized least-squares problem, typically involving the $\ell^1$-norm.
    \item In \textit{learning-based} methods, one makes little or no explicit assumptions about $\mathcal{M}_1$. Instead, one is given a \textit{training set} $\mathcal{T} = \{(y_i,x_i)\}^{K}_{i=1} \subset \C^N$, where $y_i = A x_i +e_i$ are noisy measurements of $x_i$ (the $x_i$'s are often assumed to be a subset of $\mathcal{M}_1$). One then uses $\cT$ to learn a reconstruction map $\widehat{\Psi} \colon \C^m \to \C^N$. For instance, given a class  $\cN \cN$ of NNs $\varphi \colon \bbC^m \rightarrow \bbC^N$, a regularization term $J\colon \mathcal{NN} \to \R_{\geq 0}$ and a regularization parameter $\lambda \geq 0$, a standard choice involves (approximately) solving the regularized training problem
\be{
\label{regularization}
    \min_{\varphi \in \mathcal{NN}} \frac{1}{|\cT|} \sum_{(y,x) \in \mathcal{T}} \frac12 \|\varphi(y) - x\|^2 + \lambda J(\varphi).
}
\end{enumerate}
Regardless of how it is constructed, in this paper we say that a reconstruction map $\Psi$ is \textit{stable} if small perturbations in the input $y$ yield small changes in the output $\Psi(y)$. Otherwise, we say that $\Psi$ is \textit{unstable} (or that \textit{instabilities arise}), i.e., certain small perturbations in $y$ (instabilities) cause large changes in $\Psi(y)$.

\subsection{Summary of main results}

We now present a nontechnical summary of our main results. See \S \ref{s:main-res} for the formal statements.

\begin{figure}[t]
    \centering
		\begin{tabular}{@{}>{\centering \small}m{0.3\textwidth}>{\centering \small }m{0.3\textwidth}>{\centering \small \arraybackslash}m{0.3\textwidth}@{}}   
            $x_{\mathrm{br}} + x_{\mathrm{th}} + x_{\mathrm{mi}}$ & $x_{\mathrm{br}} + x_{\mathrm{th}}$ & $x_{\mathrm{br}}$ \\ 
    \includegraphics[width=\linewidth]{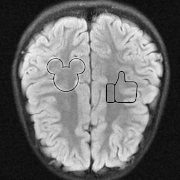} & 
    \includegraphics[width=\linewidth]{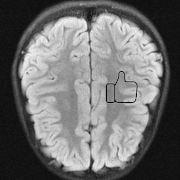} & 
    \includegraphics[width=\linewidth]{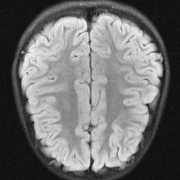}  \\ 
            $\Psi(A(x_{\mathrm{br}} + x_{\mathrm{th}} + x_{\mathrm{mi}}))$ & $\Psi(A(x_{\mathrm{br}} + x_{\mathrm{th}}))$ & $\Psi(Ax_{\mathrm{br}})$ \\ 
    \includegraphics[width=\linewidth]{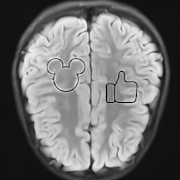} &
			\begin{overpic}[width=\linewidth]{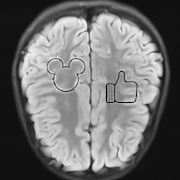}
				\linethickness{1.4pt}
				\put(24,79){\color{red}\vector(1,-2){6}}
            \end{overpic} &
			\begin{overpic}[width=\linewidth]{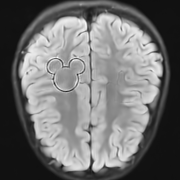}
				\linethickness{1.4pt}
				\put(24,79){\color{red}\vector(1,-2){6}}
			\end{overpic} \\
		\end{tabular}
	
    \caption{\label{fig:2}\footnotesize(\textbf{Hallucinations due to detail transfer}).
A trained NN $\Psi \colon \bbC^m \rightarrow \bbC^N$ accurately recovers the detail image in the first column. But it hallucinates by incorrectly transferring the `Mickey Mouse' detail $x_{\mathrm{mi}}$ in the first column when recovering the images in the second and third columns. The measurement matrix is a subsampled Fourier transform with $m/N = 20\%$, which models a MRI acquisition with $5$-fold acceleration. See \S \ref{s:figures-experiments} for further details. Note that the the NN does not transfer the `Thumb' detail $x_{\mathrm{th}}$. Theorem \ref{thm:AIhal2} explainns why this is the case.}
\end{figure}

\begin{mainresult}[{Hallucinations due to detail transfer -- Theorem \ref{thm:AIhal2}}]\label{m:detail} 
Let $x \in \bbC^N$ and $x_{\mathrm{Det}} \in \bbC^N$ be a detail that either belongs or lies close to $\mathcal{N}(A)$, i.e., $\tnm{A x_{\mathrm{Det}}} \ll 1 $ for some norm $\vertiii{\cdot}$. Then the following hold.
\begin{itemize}[leftmargin=0.75cm]
\item[(i)]
{Any map $\Psi$ that recovers the detail image $x + x_{\mathrm{Det}}$ will hallucinate by incorrectly transferring this detail when reconstructing the detail-free image $x$, i.e., $\Psi(A x + e) \approx x + x_{\mathrm{Det}}$. Thus, a hallucination occurs.}
\item[(ii)]
There always exist NNs {(with bounded Lipschitz constants)} that can recover details belonging to or close to $\mathcal{N}(A)$. Thus, {a NN with small error over a set of images (e.g., the training set) is liable to hallucinate.}
\end{itemize}
\end{mainresult}
The main consequence of this result is that there is an \textit{accuracy-hallucination barrier}. If the map $\Psi$ performs too well on a certain image $x_1$ with detail lying close to $\cN(A)$, then it will hallucinate by incorrectly transferring this detail to another image $x_2$. Note this situation can arise even when $\cN(A) = \{0\}$: if $A$ is ill-conditioned then there exist many `candidate' details $x_{\mathrm{Det}}$ for which $A x_{\mathrm{Det}}$ is small while $x_{\mathrm{Det}}$ is not. In Fig.\ \ref{fig:2} we present an example of this effect. A NN is trained to accurately recover a brain image with artificial details. Then when used to reconstruct the detail-free brain image, it hallucinates one of the details. Theorem \ref{thm:AIhal2} also explains why one of the details is transferred in this case, but the other is not.

Note that the reconstruction map in Main Result \ref{m:detail} can be completely stable. In other words, hallucinations are not necessarily a result of instability. As we see below in Main Result \ref{m:overperformance}, instability of a reconstruction map can also cause hallucinations, but it is not necessary prerequisite for their appearance.

Main Result \ref{m:detail} considers the performance of a reconstruction map on a single image. In our second main result, we consider a more standard training scenario where a reconstruction map is trained to perform well on a class of images.

\begin{mainresult}[{No free lunch I: overperformance implies hallucinations, yet non-hallucinating algorithms exist -- Theorem \ref{thm:AIhal1}}]
{Suppose that $\cT \subset \mathbb{C}^N$ is a finite set and $\Psi$ is a reconstruction map that achieves small error over $\cT$. Then there are infinitely many model classes $\cM_1$ with $\cT \subset \cM_1$ such that $\Psi$ hallucinates on $\cM_1$ with high probability (regardless of the distribution on $\cM_1$). However, there exists an algorithm for computing NNs that achieve small errors on $\cM_1$ and therefore do not hallucinate on $\cM_1$. }
\end{mainresult}
The main consequence of this result is that hallucinations arise necessarily as a result of overperformance of a reconstruction map that has no knowledge of the model class $\cM_1$. But if information about $\cM_1$ is given, then there are reconstruction maps -- and specifically, NN reconstruction maps -- that do not hallucinate. These NNs can also be computed in finitely many arithmetic operations and comparisons. Note that the hallucinations described by this result are \textit{in-distribution} hallucinations: namely, $\Psi(A x) \approx x +x_{\mathrm{Det}}$ for some $x \in \cM_1$ belonging to the model class. These are potentially far more problematic than \textit{out-of-distribution} hallucinations.

\begin{mainresult}[{No free lunch II: over- or inconsistent performance implies both hallucinations and instabilities -- Theorem \ref{thm:locallipcte}}]
\label{m:overperformance}
{Consider two distinct images $x$, $x'$ whose difference lies in or close to $\cN(A)$, i.e., $\tnm{A(x - x')} \ll 1$ for some norm $\tnm{\cdot}$. If $\Psi$ recovers both $x$ and $x'$ well (overperformance), or recovers $x$ well and $x'$ poorly (inconsistent performance), then the following must hold.
\begin{itemize}[leftmargin=0.75cm]
\item[(i)]
$\Psi$ is unstable in a ball around $y = A x$, with the instability becoming worse as the reconstruction performance improves.
\item[(ii)]
$\Psi$ hallucinates in a ball around $y = A x$: there are small perturbations $e$ for which, when given measurements $A x + e$, $\Psi$ produces false details not in the image $x$.
\end{itemize}
}
\end{mainresult}
A key consequence of this result is that there is an \textit{accuracy-stability trade-off}. If the reconstruction map $\Psi$ overperforms (on as little as two images) then it is necessarily unstable, with the instability becoming arbitrarily large as the performance increases. This result also asserts that instabilities and hallucinations are \textit{stable}. Bad perturbations do not belong to a set of Lebesgue measure zero. In fact, there are balls of such perturbations. See also Main Result \ref{m:notrare} below.

Put another way, Main Result \ref{m:overperformance} states that a reconstruction map can only be stable if it does not overperform on certain images. This is a problem for learning-based algorithms, which are typically trained to achieve small error over the training set. According to this result, if there are two images $x_1,x_2$ in the training set whose difference $x_1 - x_2$ has small measurements $A(x_1-x_2)$, then the learned map will necessarily be unstable.

For this reason, successful training should balance stability and performance. However, as we discuss later in \S \ref{ss:no-free-II} and Fig.\ \ref{fig:3}, devising training strategies that optimize the trade-off between performance and stability is by no means straightforward.

Main Result \ref{m:overperformance} asserts the existence of `bad' perturbations that cause either instabilities or hallucinations. A natural question to ask is whether these are rare events. In the next main result, we show that they not need to be.

\begin{figure}[t]
\begin{center}
    \setlength{\tabcolsep}{2pt}
    \begin{tabular}{@{}l@{}}
    \begin{tabular}{@{}>{\centering}m{0.48\textwidth}>{\centering\arraybackslash}m{0.48\textwidth}@{}}
         Hallucinations with zero-mean Gaussian noise &  Instabilities w.r.t. to Gaussian noise with image independent mean \\
        \begin{tabular}{@{}>{\small \centering}m{0.235\textwidth}>{\small \centering\arraybackslash}m{0.235\textwidth}@{}}
             Noisy image: & Noisy image: \\
             $|x+v|$ & $|x+v|$ \\ 
             (full size)& (cropped) \\
       \includegraphics[width=\linewidth]{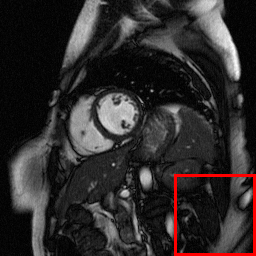}
     & \includegraphics[width=\linewidth]{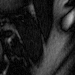} \\
            DeepMRI-net: & DeepMRI-net: \\
            $|\Psi(A(x+v))|$ & $|\Psi(Ax)|$ \\
            (cropped) & (cropped) \\
            \begin{overpic}[width=\linewidth]{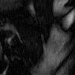}
			\linethickness{2.4pt}
            \put(68,42){\color{red}\vector(1,4){6}}
		\end{overpic}
       &\includegraphics[width=\linewidth]{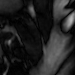}
    \end{tabular}
   &\begin{tabular}{@{}>{\small \centering}m{0.235\textwidth}>{\small \centering\arraybackslash}m{0.235\textwidth}@{}}
       AUTOMAP: &  AUTOMAP:  \\ 
        $\Psi(Ax+e_0)$ & $\Psi(Ax+e_1)$ \\
             (full size)& (full size) \\
        \includegraphics[width=\linewidth]{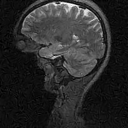} 
       &\includegraphics[width=\linewidth]{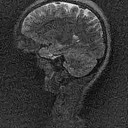} \\
       AUTOMAP: &  AUTOMAP:  \\ 
        $\Psi(Ax+e_2)$ & $\Psi(Ax+e_3)$ \\
             (full size)& (full size) \\
        \includegraphics[width=\linewidth]{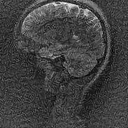}
       &\includegraphics[width=\linewidth]{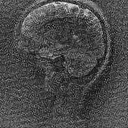}\\
    \end{tabular} \\
    
    \end{tabular} 
    \end{tabular}
\end{center}
\caption{\label{fig:deep_mri_pert12}(\textbf{Hallucinations and instabilities due to random noise})
    Two DL methods exhibit hallucinations and instabilities due to random noise. On the left, the DeepMRI-net \cite{SchCaH-17} reconstruction map is unstable to mean-zero Gaussian noise $v$. In this case, the NN hallucinates by removing a key image feature (see the red arrow). On the right, the AUTOMAP \cite{Bo-18} reconstruction map is unstable to Gaussian noise. The noise vector $e_0$ is drawn from a zero-mean Gaussian distribution, whereas the mean of the distribution,  used to generate $e_1,e_2$ and $e_3$, is based on three worst-case noise vectors computed for AUTOMAP with respect to a different image. This makes the mean is image independent. The instability of the reconstruction map produces noticeable artefacts in all cases. The measurement matrix in these experiments is a subsampled Fourier transform with 33\% (left) and 60\% (right) subsampling, respectively. See \S \ref{s:figures-experiments} for further information.
}

\end{figure}

\begin{mainresult}[Instabilities and hallucinations are not rare events -- Theorems \ref{thm:locallipcte-random} and \ref{thm:worstvar}]
\label{m:notrare}
Consider the same conditions as Main Result \ref{m:overperformance} and let $E$ be an absolutely continuous random vector with a strictly positive probability density function. Then, with nonzero probability,
\begin{itemize}[leftmargin=0.75cm]
\item[(i)] $\Psi$ is unstable, i.e., $\nm{\Psi(A x + E) - \Psi(A x)} \gg 0$,
\item[(ii)] $\Psi$ hallucinates, i.e., when applied to measurements $A x + E$, it produces false details not in the image $x$.
\end{itemize}
Moreover, for any $0 < \delta < 1$ there is a Gaussian distribution with small mean for which this holds with probability at least $1-\delta$. And subject to several additional conditions, the variance of this Gaussian distribution tends to zero as $m \rightarrow \infty$.
\end{mainresult}
This result implies that random noise can also produce undesirable effects. In Fig.\ \ref{fig:deep_mri_pert12} we show several examples of this effect. For the first DL method, mean zero Gaussian noise causes the NN reconstruction map to hallucinate, by artificially removing an image feature (indicated by the red arrow). In the second case, certain \textit{image independent}, \textit{small mean} Gaussian noise causes severe instabilities in the recovered image. Notice that the noise causes the second DL method to exhibit completely nonphysical artefacts, which could be easily identified by a practitioner as a failure mode. Yet the first DL method creates seemingly realistic artefacts (i.e., hallucinations). Such pernicious artefacts may be impossible to detect.

Having considered hallucinations and instabilities, in the next result we switch focus and consider the performance of reconstruction maps in terms of their accuracy (generalization). We do this by following the well-established framework of \textit{optimal recovery} (or, as we term it, \textit{optimal maps}). Specifically, we consider the existence of mappings $\Psi \colon \cM_2 = A\cM_1 \rightarrow \bbC^N$ that achieve the smallest possible worst-case error over a given model class $\cM_1$.  This is a topic with a long history \cite{micchelli1977survey}, but one that has been the subject of renewed interest in recent years. See the seminal work of Cohen, Dahmen \& DeVore \cite{CDDCSkterm} and, more recently, \cite{binev2022optimal, bourrier2014fundamental} and references therein.

\begin{mainresult}[Optimal maps may be impossible to train -- Theorem \ref{thrm:not_optimal}]
Let $\delta > 0$ be sufficiently small. Then the following hold.
\begin{itemize}[leftmargin=0.75cm]
\item[(i)] There are uncountably many model classes $\cM_1$ and sets $\cT$ for which any Lipschitz continuous map $\Psi$ that produces an error of at most $\delta$ over $\cT$ cannot be an optimal map (or an approximately optimal map) over $\cM_1$.
\item[(ii)] There are uncountably many model classes $\cM_1$ for which no Lipschitz continuous map can attain an error of $\delta$ over $\cM_1$.
\end{itemize}
\end{mainresult}
Part (i) of this result asserts that training may fail to yield optimal maps. A small error over $\cT$ (which may, for example, correspond to the training set, or the union of the training and test sets) offers no guarantee that the learned map be optimal over the model class. Put another way, the learned map may suffer from \textit{inconsistent} performance. Part (ii) asserts that there are model classes $\cM_1$ for which there is no map that can achieve a small error over $\cM_1$. In other words, the map (implicitly) sought by training may not exist in the first place.

\subsection{Related work}\label{s:relwork} 

We conclude this section with a discussion of related work.

\subsubsection{Traditional and AI-based methods for image reconstruction}

Inverse problems in imaging is a vast topic, with a history dating back many decades and encompassing many different methodologies. Traditional image reconstruction methods are typically model-based. They try to solve \eqref{eq:invintro2}, given an accurate description of $A$ and $\mathcal{M}_1$. Here $\mathcal{M}_1$ could be the subspace $\mathcal{N}(A)^{\perp}$, or the space of images which are (approximately) sparse in a wavelet or discrete gradient transform (or their various generalizations). In cases where $A$ is poorly conditioned, Tikhonov regularization has been widely used. However, with the advent of compressive sensing in the mid-2000s, sparsity promoting methods using different types of $\ell^1$-regularization gained popularity. These methods allowed practitioners to reduce sampling rates beyond those of standard methods at the time, while preserving reconstruction accuracy. Today, DL-based methods have surpassed the accuracy of all traditional methods. However, recent results have also indicated that traditional methods fine-tuned by using ML can achieve comparable accuracy to DL-based methods  for moderate acceleration factors \cite{gu2022revisiting}. For recent overviews of traditional methods and more on the transition to data-driven methods, we refer the reader to \cite{adcock2021compressive,  mccann2019biomedical, ravishankar2019image} and references therein. For an overview of the progress of AI- and DL-based methods, we refer the reader to some of the many recent review articles on this topic \cite{ArridgeEtAlACTA, liang2020deep,  Mccannetal-17, ongie2020deep, ravishankar2019image, DL_for_PET_review, DL_for_CT_review, new_frontier18}.

\subsubsection{Instabilities, generalization and hallucinations in AI}
\label{sss:Instabilities}

The issue of instabilities in DL has been an active area of inquiry ever since Szegedy et al. \cite{SzZ-14} demonstrated that state-of-the-art deep NNs used in image classification are unstable to certain small adversarial perturbations of the input. By now, the phenomenon seems ever present \cite{heaven2019deep, NNInstab} in most state-of-the-art DL techniques used in, e.g., image classification \cite{eykholt2018robust}, audio and speech recognition  \cite{Carlini18}, natural language processing and automated diagnosis in medicine \cite{Science_adv}. 

  For image reconstruction, the issue of instabilities for DL methods was first investigated empirically in \cite{Anetal-19, huang18}. This was followed by further investigations  on the phenomenon in \cite{Heckel21, genzel2020solving}, which also included stability tests for sparse regularization decoders and investigations of distribution shifts for both DL and sparse regularization methods. A key finding in these works is that both classes of methods are susceptible to worst-case noise and distribution shifts in the data (recall Remark \ref{rem:sparse-reg-also}). However, other results have also suggested robustness of sparse regularization to worse-case noise \cite{colbrook22,neyra-nesterenko2023stable},\cite[Chpt.\ 21]{adcock2021compressive}. Thus, different implementations of sparse regularization techniques may affect their reliability. As was also noted in \cite{zhang2021instabilities}, the instabilities observed in \cite{Heckel21} for parallel MRI may also be influenced by the moderate ill-conditioning of the forward problem for high acceleration factors. 

The strategies in \cite{Anetal-19, Heckel21, huang18, genzel2020solving} are based on finding a worst-case perturbation in the $\ell^2$-norm. Other `attack' strategies have also been proposed. Examples include using NNs to generate the worst-case noise \cite{bresler20}, localized attacks \cite{alaifari2022localized, morshuis2022adversarial} or attacks based on rotation \cite{morshuis2022adversarial}. As noted, \cite{genzel2020solving} also empirically studied the accuracy-stability trade-off for DL in imaging (recall Fig.\ \ref{fig:3}).
  
        In \cite{Anetal-19} it was observed that NNs can be unstable to changing acquisition strategies. In particular, acquiring more samples does not necessarily enhance accuracy when using NNs, unlike with standard methods. This behaviour has been further investigated in \cite{gilton2021model, weiss22adaptive, johnson2021evaluation}. In \cite{gilton2021model} several strategies were developed to tackle this problem. In \cite{weiss22adaptive}, the authors demonstrate empirically that training on multiple sampling patterns in MRI can improve robustness towards changing acquisition strategies, while in \cite{johnson2021evaluation} the authoes test the submissions from the 2019 fastMRI challenge \cite{fastmri19} with respect to changing acquisition settings. 

            The issue of robustness toward changing acquisition strategies is closely related to the issue of distribution shifts in ML. That is, how does a method trained on one dataset generalize to a different dataset? This question has received increasing attention in imaging \cite{lustig_pnas22, yu2022validation, Heckel21, fastmri20}, as DL-based methods are now starting to enter clinical practice \cite{GE_DL_technology, Philips_DL_technology}. In \cite{lustig_pnas22} the authors study how common ML training pipelines can lead to models that are biased towards the training data, whereas in \cite{yu2022validation} the authors investigates the generalizability of self-supervised methods for both prospective and retrospective MRI data. In \cite{Heckel21} the authors investigate distribution shifts for learning and model-based methods. The 2020 fastMRI challenge also introduced a test to check a NN's ability to generalize on data from different MRI scanners \cite{fastmri20}. 

        As highlighted in \S \ref{sec:introduction}, the issue of AI-generated hallucinations is raising concerns within the imaging community. Hallucinations have the potential to cause misdiagnoses in medicine \cite{fastmri19, fastmri20, Bha20} and to hinder discoveries in the life sciences \cite{bel19, hoff21}. To simplify research on hallucinations in medical imaging, the fastMRI+ \cite{fastmri+} and SKM-TEA \cite{desai2022skm} datasets have recently been introduced. These datasets contain expert annotations of relevant pathologies and thereby allow practitioners to automate the search for hallucinations in reconstructed images. It has also become customary to test different models' ability to reconstruct unseen features in the images, by adding small details not contained in the training dataset, see e.g., \cite[Fig.\ 2]{johnson2021evaluation} and \cite[Fig.\ 12]{ongie2020deep}. This test was initially proposed in \cite{Anetal-19}.

        It is also worth noting that several generative models have been shown to hallucinate. In the landmark publication \cite{bora2017compressed}, one can, for example, see in Fig.\ 15c how the generative model puts red lipstick on a man's lips and draws eyes on top of sunglasses. Furthermore, in \cite[Fig.\ 3]{jalal2020robust} one can see how different generative reconstruction algorithms produce nonsensical, but realistic-looking faces when the matrix $A$ and measurements $y$ are corrupted by noise. While the issue of hallucinations is not the focus of these works, it shows that this phenomenon is pervasive.

\section{Preliminaries}
Before stating our main results, we require some additional notation.
Given a set $\mathcal{M}_1 \subset \C^N$ and a matrix $A \in \C^{m \times N}$, we let $\mathcal{M}_2 = A\mathcal{M}_1 = \{Ax : x \in \mathcal{M}_1\}$. 
For a set $\Omega \subset \{1,\ldots, N\}$, we let $P_{\Omega} \in \C^{N\times N}$ denote the projection onto the canonical basis indexed by $\Omega$, i.e., for $x \in \C^{N}$, $(P_{\Omega}x)_i = x_i$ if $i\in \Omega$ and $0$ otherwise. We sometimes abuse notation slightly and assume that $P_{\Omega} \in \C^{m\times N}$, where $m=|\Omega|$, by ignoring the zero entries. 

Throughout we let $\nm{\cdot}$ denote a norm on $\C^N$ and $\vertiii{\cdot}$ denote a norm on $\C^m$. We let $\mathcal{B}(x, r) = \{z \in \C^N : \|x-z\| \leq r\}$ denote the closed ball centered at $x \in \C^N$ with radius $r > 0$. If $y\in \C^{m}$, then we write $\mathcal{B}(y,r) \subset \bbC^m$ for the ball centred at $y$ with radius $r$ in the norm $\vertiii{\cdot}$. For a set $\mathcal{M}_1 \subset \C^N$, we write
\[
\mathcal{M}_1^{\nu} = \{z \in \mathbb{C}^N : \exists x \in \mathcal{M}_1 \text{ such that } \|z-x\| < \nu \}
\] 
for the $\nu$-neighborhood of $\mathcal{M}_1$ in $\C^N$. We define $\mathcal{M}_2^{\nu}$ for $\cM_2 \subset \bbC^m$ in the same way. 
For $\epsilon > 0$ and $y \in \bbC^m$, we define the local $\epsilon$-Lipschitz constant and global Lipschitz constant of a mapping $\Phi \colon \bbC^m \rightarrow \bbC^N$ as, respectively,
\[
L^{\eps}(\Phi,y) = \sup_{\substack{z \in \C^m \\ 0 < \vertiii{z-y} \le \eps}} \frac{\|\Phi(z)-\Phi(y)\|}{\vertiii{z-y}} \quad \text{and}\quad 
L(\Phi) = \sup_{\substack{y,z \in \bbC^m \\ z \neq y }} \frac{\nm{\Phi(z) - \Phi(y)}}{\vertiii{z-y}}
\]
Several of the theoretical results in this paper pertain to NN reconstruction maps. For the purposes of this paper, a NN is any map $\Psi : \bbC^m \rightarrow \bbC^N$ of the form
\be{
\label{Psi-form}
\Psi(y) = W \Phi(y) + b
}
 for some map $\Phi : \bbC^m \rightarrow \bbC^n$, weight matrix $W \in \bbC^{N \times n}$ and bias $b \in \bbC^N$. A \textit{class} or \textit{family} of NNs is a collection of the form
\be{
\label{NN-family}
\cN\cN = \{ y \mapsto W \Phi (y) + b : W \in \bbC^{N \times n},\ b \in \bbC^N, \Phi \in \cF \},
}
where $\cF$ is a family of maps $\Phi : \bbC^m \rightarrow \bbC^n$.
This definition is extremely general, and includes most types of NN constructions used in practice (as well as many constructions that would not be considered NNs). In particular, it includes any NN architecture in which the output layer is not subject to a nonlinear activation, as common in regression problems. As a result, this means that the conclusions of our main theorems are very generally applicable.

For many imaging modalities, the noise is modelled as a complex random variable drawn from a complex normal distribution. We denote this by $\C\mathcal{N}(v, \Sigma)$, where $v\in \C^{m}$ is the mean and $\Sigma \in \C^{m\times m}$ is the positive semi-definite covariance matrix. We note that the probability density function of $\C\mathcal{N}(v, \Sigma)$ exists if $\Sigma$ is positive definite  and is given by 
\[ \pi^{-m}\det(\Sigma)^{-1} \exp(-(u-v)^*\Sigma^{-1}(u-v)), \quad u \in \C^m.  \]
See \cite{complex_gauss_book} and, in particular, \cite[Thm.\ 2.10]{complex_gauss_book}.

\section{Main results}\label{s:main-res}

We now present our main results. Proofs are deferred to \S \ref{s:dismain}.

\subsection{Hallucinations due to detail transfer}

Recall that a \textit{hallucination} is a realistic-looking, but ultimately false detail arising in the reconstruction of a given image. There are various mechanism that can cause hallucinations, one of which is \emph{detail transfer}. Detail transfer means that a detail from one image -- typically, in the case of DL, this is an element of the training set from which the given NN reconstruction map is learned -- is transferred to another image via the reconstruction map. See Fig.\ \ref{fig:2} for an illustration of this process. In the following theorem, we provide a mathematical explanation for hallucinations arising via detail transfer.

\begin{theorem}\label{thm:AIhal2}
    Let $A \in \bbC^{m \times N}$, {$\delta,L > 0$ and $x, x_{\mathrm{Det}} \in \mathbb{C}^N$ with $\vertiii{Ax_{\mathrm{Det}}} \leq \delta$.  }
\begin{enumerate}[leftmargin=0.75cm,label=(\roman*)]
\item \emph{($\Psi$ hallucinates by transferring details).}\ {Let $\Psi \colon \C^{m}\to\C^{N}$ be Lipschitz continuous with constant at most $L$ and suppose that}
	\begin{equation}\label{eq:NN_exists}
		\|\Psi\big(A(x+x_{\mathrm{Det}})\big) - (x+x_{\mathrm{Det}})\| \leq \delta.
	\end{equation}
	Then for every $e \in \mathcal{B}(0,\delta)$, there is a $z \in \mathbb{C}^N$ with $\|z\| \leq (1+ 2L)\delta$, such that
	\be{
\label{eq:detail-transfer}
	\Psi(Ax + e) = x + x_{\mathrm{Det}} + z.  
}
\item	\emph{(There always exists a NN that hallucinates)}.\ For any family $\cN\cN$ of the form \eqref{NN-family}, there is a $\widetilde{\Psi} \in \cN\cN$ with Lipschitz constant at most $L$ that satisfies \eqref{eq:NN_exists}.
\end{enumerate}	
\end{theorem}
Let $\delta > 0$ be some small number. Then this theorem asserts that the detail $x_{\mathrm{Det}}$ in the image $x' \coloneqq x + x_{\mathrm{Det}}$ will be transferred onto the detail-free image $x$ given noisy (or noiseless) measurements $A x+e$, with $e\in \mathcal{B}(0,\delta)$. 
    
This theorem does not require a formal definition of what constitutes a detail. Informally, a detail $x_{\mathrm{Det}}$ should also be a `local' feature, whose presence or absence is clearly visible and relevant to the given problem. A `global' feature, such as a texture, may not be noticeable, even if relatively large in norm. Regardless, this theorem becomes relevant when $\nm{x_{\mathrm{Det}}} \gg \delta \geq \tnm{A x_{\mathrm{Det}}}$. In other words, the detail is significant (large in norm), but has small measurements. As noted, this situation can readily occur when $A$ has a nontrivial kernel $\cN(A)$ (i.e., $\mathrm{rank}(A) < N$), as is common in many applications (e.g., when $m < N$). But it may also arise when $\cN(A) = \{ 0\}$ (i.e., $\mathrm{rank}(A) = N$) but $A$ is ill-conditioned, as is common in others.

\begin{figure}[t]
    \centering
	\begin{tabular}{@{}>{\centering \small }m{0.3\textwidth}>{\centering \small \arraybackslash}m{0.3\textwidth}@{}}
        $x + x_{\mathrm{Det}}$ & $x$  \\
        \includegraphics[width=\linewidth]{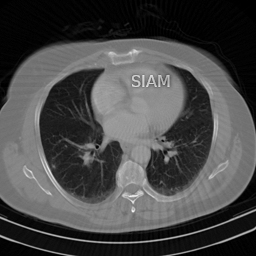} & 
        \includegraphics[width=\linewidth]{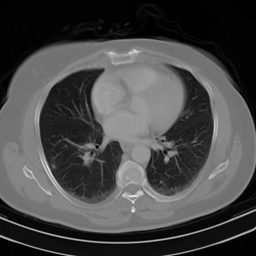}  \\ 
        $\Psi (A(x + x_{\mathrm{Det}}))$ & $\Psi (Ax)$ \\ 
        \includegraphics[width=\linewidth]{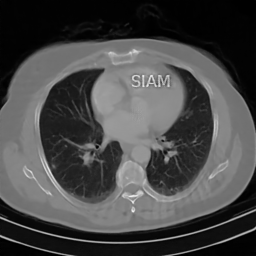} & 
        \begin{overpic}[width=\linewidth]{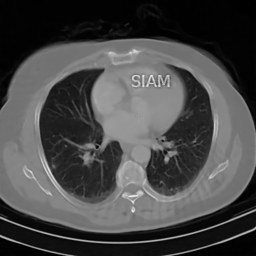}
        	\linethickness{1.4pt}
        	\put(45,82){\color{red}\vector(1,-2){6}}
        \end{overpic} \\
    \end{tabular}
    \caption{\label{fig:new_fig}(\textbf{Hallucinations due to detail transfer}). 
A NN $\Psi\colon \R^{m}\to \R^N$ is trained to accurately recover CT images and the image $x+x_{\mathrm{Det}}$ with the \enquote{SIAM} detail inserted, as shown in the first column. This causes the NN to incorrectly transfer the \enquote{SIAM} detail when reconstructing the detail-free image $x$ from measurements $Ax$, just as described by Theorem \ref{thm:AIhal2}. In this experiment the matrix $A$ is a subsampled Radon transform which samples 50 equally spaced angles, and the detail $x_{\mathrm{Det}}$ is designed such that $0 < \|Ax_{\mathrm{Det}}\|_{\ell^2} \ll \|Ax\|_{\ell^2}$.  
}
\end{figure}

Fig.\ \ref{fig:2} presents an example of this result. In this figure, the `Mickey Mouse' detail $x_{\mathrm{mi}} \in \cN(A)$, whereas the `Thumb' detail $x_{\mathrm{th}}$ has relatively large measurements, i.e., $\tnm{A x_{\mathrm{th}}} \gg 0$. The NN is trained to recover the image $x_{\mathrm{br}} + x_{\mathrm{th}} + x_{\mathrm{mi}}$. As a result, it incorrectly transfers the detail $x_{\mathrm{mi}}$, while the detail $x_{\mathrm{th}}$ is handled correctly (i.e., it is not transferred). Fig.\ \ref{fig:new_fig} shows another example of this effect. In this case, $A$ is a subsampled Radon transform, which models a CT imaging scenario. Here, the NN is trained to recover the detail image $x + x_{\mathrm{Det}}$, and as a result, it incorrectly transfers the detail $x_{\mathrm{Det}}$ when recovering the detail-free image $x$.

Observe that Theorem \ref{thm:AIhal2} does not require the map $\Psi$ to be unstable. It arises because detail $x_{\mathrm{Det}}$ lies close to the kernel of $A$, i.e., $\vertiii{Ax_{\mathrm{Det}}}\leq \delta$, and the reconstruction map $\Psi$ recovers $x'$ well. Since the measurements $y' = A x'$ and $y = A x + e$ are similar, i.e., $\nm{y-y'} \leq 2 \delta$, the Lipschitz continuity of $\Psi$ means that it must, when presented with measurements $y$, produce an image that is close to $x'$.

Theorem \ref{thm:AIhal2} also shows how alarmingly easy it is for a reconstruction map to hallucinate. Simply recovering an image with a detail $x_{\mathrm{Det}}$ lying close to $\cN(A)$ is enough for hallucinations to arise. As mentioned, measurement matrices arising in imaging problems often have large kernels. For example, in undersampled MRI one may typically consider images of size $N = 256 \times 256 = 65,536$ and a subsampling factor of $25\%$, giving $m = 65,536 / 4 = 16,384$ measurements. Hence $\cN(A)$ has dimension $N - m = 49,152$. Therefore, given a training set of typical MRI images, the large dimension of $\cN(A)$ means there may well be many possible ways in which the conditions that lead to part (i) of Theorem \ref{thm:AIhal2} can arise.

Part (ii) of Theorem \ref{thm:AIhal2} implies that hallucinations due to detail transfer can occur very easily in, for instance, DL. If a NN is trained using a training set that contains images with details $x_{\mathrm{Det}}$ -- for instance, small tumours or lesions in the case of medical imaging -- that have small measurements $A x_{\mathrm{Det}}$, then it will be liable to hallucinate such details when subsequently applied to images outside of the training set that lack this detail. Put another way, $\Psi$ may produce \textit{false positives}. Or, if the scenario is reversed, it may produce \textit{false negatives}. Both effects are highly undesirable.

\begin{remark}
\label{rem:why-not-CS}
It is natural to ask whether model-based methods are also susceptible to the conclusions of Theorems \ref{thm:AIhal2}. Typically they are not, provided $A$ satisfies appropriate conditions that ensure good performance of the model-based reconstruction map over the chosen model class $\cM_1$. We explain this further in \S \ref{s:model-based-CS} in the case of sparse regularization based on $\ell^1$-minimization, where $\cM_1$ is the set of approximately sparse vectors in some unitary transform and $A$ is assumed to satisfy the \textit{robust Null Space Property (rNSP)}. The theory of compressed sensing then militates against the appearance of hallucinations of the type described by Theorem \ref{thm:AIhal2}.  
\end{remark}

\subsection{No free lunch I: overperformance implies hallucinations, yet non-hallucinating algorithms exist}

Since the previous theorem is very general, the question arises as to how it relates to learning-based methods -- i.e., those that use a training set to learn a reconstruction map -- for solving inverse problems. The following theorem elaborates on this relation. 

\begin{theorem}\label{thm:AIhal1}
    Let $A \in \C^{m\times N}$, $\mathcal{T} \subset \mathbb{C}^N$ be a non-empty and finite set, $\delta > 0$, $\cN\cN$ be a family of NNs as in \eqref{NN-family}, $\Psi \in \cN \cN$ have Lipschitz constant $L > 0$ and $x_{\mathrm{Det}} \in \C^N$ with $\vertiii{Ax_{\mathrm{Det}}} \leq \delta/(4L)$. 
     Suppose that $\Psi$ satisfies
\[
\max_{x \in \mathcal{T}} \|\Psi(Ax) - x\| \leq \delta.
\]
     Then, for any $\epsilon \geq \delta/(2L)$ there is an uncountable family $\mathcal{C}$ of  finite or countably infinite sets 
     $\mathcal{M}_1 \subset \mathbb{C}^N$ with $ \mathcal{T} \subset \mathcal{M}_1$ and $A\mathcal{M}_1 \subset (A\mathcal{T})^{\epsilon}$, such that for each $\mathcal{M}_1 \in \mathcal{C}$ the following hold simultaneously.
\begin{enumerate}[leftmargin=0.75cm,label=(\roman*)]
		\item \emph{($\Psi$ suffers from in-distribution hallucinations).} 
		For any probability distribution $\mathcal{D}$ on $\mathcal{M}_1$ with the property that $\mathbb{P}_{X \sim \cD}(X \in \mathcal{T}) \leq q$ it holds that
		\begin{equation*} 
            \mathbb{P}_{X \sim \cD}\Big(\exists \lambda \in \C, |\lambda|=1 \text{ such that } \|\Psi(AX) - (X + \lambda  x_{\mathrm{Det}}) \| \leq 2\delta \Big) \geq 1-q.
		\end{equation*}
		\item \emph{(There exists an algorithm that yields non-hallucinating NNs).}
            There exists an algorithm $\Gamma : \bbC^m \rightarrow \cN \cN$ such that, for all $x \in \cM_1$,
\[
\|\Phi_{Ax}(Ax) - x\| \leq \delta,\quad \text{where }\Phi_{Ax} = \Gamma(A x) \in \cN \cN.
\]
	\end{enumerate}
\end{theorem}
As mentioned after Theorem \ref{thm:AIhal2}, the assumption  $\vertiii{Ax_{\mathrm{Det}}} \leq \delta/(4L)$ can easily occur when $A$ is ill-conditioned or when $A$ has a nontrivial kernel. We also note that the assumption that $\Psi$ is a NN is only needed for part (ii) of the statement. Part (i) holds for any Lipschitz continuous reconstruction map $\Psi \colon \bbC^m \rightarrow \bbC^N$.
Part (i) of this theorem therefore says that a reconstruction map that \textit{overperforms} on a set $\cT$ (e.g., a training set) that belongs to a model class $\cM_1$ must, with high probability, hallucinate over $\cM_1$. We refer to these as \textit{in-distribution} hallucinations, since the inputs $X$ which suffer from hallucinations are drawn from a distribution over $\cM_1$. Recall that a model class $\cM_1$ represents the images we are interested in recovering in a given application. For example, these may be brain images in MRI. Thus, while \textit{out-of-distribution} hallucinations may not be as problematic in practice, in-distribution hallucinations are far more worrying.

Part (i) of Theorem \ref{thm:AIhal1} is general in that it holds for any distribution $\cD$, with the only assumption being that $\bbP_{X \sim \cD}(X \in \cT) \leq q$. In other words, the set $\cT$ cannot have too large a measure relative to $\cM_1$ and $\cD$. The main consequence of this result is that hallucinations of this type cannot be avoided when using standard NN training strategies that simply strive for small error over a training set. Put another way, to prevent against such hallucinations, it is necessary to change the training strategy to incorporate additional information about the problem.

This brings us to part (ii). It states that there is an algorithm for computing NN reconstruction maps that achieve hallucination-free performance over $\cM_1$. Note that by `algorithm' we mean that $\Gamma$ takes a finite set of real numbers as its input (i.e., the vector $y$) and performs only finitely many arithmetic operations and comparisons, producing a finite set of real numbers as its output (i.e., the weights and biases of a NN). More formally, $\Gamma$ is a BSS (Blum-Shub-Smale) machine -- see the proof of Theorem \ref{thm:AIhal1} for details.
The key point is that, as opposed to a single NN $\Psi$ as in part (i), the NN $\Phi_y$ in part (ii) depends on the input. Note that this idea is not far-fetched. In fact, NNs that arise from unrolling optimization algorithms \cite{monga2021algorithm}, \cite[Chpt.\ 21]{adcock2021compressive} and NNs based on the deep image prior framework \cite{Ulyanov18} are generally of this type.

Finally, we remark in passing that Remark \ref{rem:why-not-CS} also applies to Theorem \ref{thm:AIhal1}.

\begin{remark}\label{not-overfitting}
It is important observe that  Theorem \ref{thm:AIhal1} is \textit{not} a statement about overfitting.  Overfitting in DL occurs when a NN performs well on the training set, but poorly on the test set.  This phenomenon is caused by the fact that the architecture of the network is fixed, and hence its ability to fit data is limited (it can fit the training set, but not the test set). It is a classical result in approximation theory that any set of data points (e.g.\ the union of the training and test sets) can be interpolated by a NN of sufficient size (see, e.g., \cite[Chpt.\ 18]{adcock2021compressive}).  So even if the trained network would suffer from overfitting, and hence lack performance on the test set, there will exist another NN that interpolates all data points in the training set as well as the test set.  What Theorem \ref{thm:AIhal1} describes is a phenomenon that happens \emph{for all} mappings. There is no restriction in the network architecture, and, in fact, in part (i) $\Psi$ need not be a NN in the first place. More directly, one could simply let the set $\cT$ in part (i) contain both the training sets and test sets. Theorem \ref{thm:AIhal1} then says that one can have excellent performance on these sets but still suffer from in-distribution hallucinations.
\end{remark}

\subsection{{No free lunch II: over- or inconsistent performance implies both hallucinations and instabilities}}\label{ss:no-free-II}

In the previous results, we described how overperformance can cause hallucinations. We now describe a second key mechanism that can cause a reconstruction map to perform in a substandard way. In the following result, we show that \textit{over- or inconsistent performance} of a reconstruction map causes both instabilities and hallucinations.

\begin{theorem}\label{thm:locallipcte}
	Let $A \in \bbC^{m \times N}$, $x,x' \in \mathbb{C}^N$ and $\eta >0$.  Let $\Psi \colon \mathbb{C}^m \rightarrow  \mathbb{C}^N$ be continuous and suppose that
	\begin{equation}\label{eq:cond_instability}
         \|\Psi(A x) - x\| < \eta,\quad\quad \|\Psi(A x') - x''\| < \eta,
	\end{equation}
	{for some $x'' \in \bbC^N$} and that
	\begin{equation}\label{eq:cond_instability2}
		\vertiii{A(x -  x')} \leq \eta.
	\end{equation}
	Then the following hold.
	\begin{itemize}[leftmargin=0.75cm]
        \item[(i)]  \emph{($\Psi$ is unstable)}.\
		There is a closed non-empty ball $\mathcal{B} \subset \mathbb{C}^m$ centred at $y = Ax$ such that, for all $\eps \geq \eta$, the local $\eps$-Lipschitz constant at any $\tilde y \in \mathcal{B}$ satisfies
            \begin{equation}\label{eq:Lipchitz}
L^{\eps}(\Psi,\tilde y) \geq \frac{1}{\eta}\left(\|x-x''\| - 2\eta \right).
\end{equation}
\item[(ii)] \emph{($\Psi$ hallucinates)}.\
    There exist $z \in \mathbb{C}^N$ with $\|z\| \geq \|x - x''\|$ {(for example, $z = x'' - x$)}, $e \in \mathbb{C}^m$ with $\vertiii{e} \leq \eta$, and closed non-empty balls $\cB_x$, $\cB_e$ and $\cB_z$ centred at $x$, $e$ and $z$, respectively, such that 
\begin{equation}\label{eq:false_pos}
\|\Psi(A \tilde x + \tilde e)- (\tilde x + \tilde z)\| \leq \eta,\quad \forall \, \tilde x \in \cB_x, \tilde e \in \cB_e,\tilde{z} \in \cB_z.
\end{equation}
	\end{itemize}
\end{theorem}
Suppose first that $x'' = x'$, so that the map $\Psi$ reconstructs both $x$ and $x'$ to within an accuracy of $\eta$. Suppose also that $x$ and $x'$ are sufficiently distinct, i.e., $\nm{x - x'} > 2 \eta$. Then \eqref{eq:cond_instability}-\eqref{eq:cond_instability2} state that $\Psi$ \textit{overperforms} in the sense that it recovers two vectors $x,x'$ well whose difference has small measurements $A (x-x')$. In other words, $\Psi$ strives to get something from nothing: the $x$ and $x'$ are distinct, but their measurements (the inputs to $\Psi$) are similar. Unsurprisingly, the result is instability, with the local Lipschitz constant in a ball around $y = A x$ scaling like $1/\eta$. And the more the reconstruction map overperforms, the worse this instability becomes.

Now suppose that $x'$ is close to $x$, but $\nm{x - x''} > 2 \eta$. In this case, $\Psi$ performs \textit{inconsistently}: it recovers $x$ well, but recovers a nearby $x'$ poorly. The conclusion is once again the same. The map $\Psi$ is necessarily unstable in a ball around $y = A x$.

\begin{figure}[t]
	\centering
	\setlength{\tabcolsep}{1pt} 
	\begin{small}
		\begin{tabular}{@{}c@{}}
			\begin{tikzpicture}
				\draw[ultra thick, {stealth[scale=3]}-{stealth[scale=3]}] (0,0) node[above right]{\parbox[b]{0.28\textwidth}{Increased accuracy \\ decreased stability}} -- 
				(0.9\linewidth,0) node[above left] {\parbox[b]{0.22\textwidth}{Increased stability\\ decreased accuracy}};
				\draw (0.45\linewidth, 0) node[above] {\parbox[b]{0.25\textwidth}{\begin{center}\textbf{Accuracy-stability} \\ \textbf{trade-off}\end{center}}};
			\end{tikzpicture}\\
			\begin{tabular}{@{}>{\centering}m{0.32\textwidth}>{\centering}m{0.32\textwidth}>{\centering\arraybackslash}m{0.32\textwidth}@{}}   
                Trained on & Trained on {moderately} & Trained on {highly} \\
                noiseless measurements  &  noisy measurements &  noisy measurements\\
				(ItNet w/o noise.) & (ItNet w/\ low noise) & (ItNet w/\ high noise) \\
				\includegraphics[width=\linewidth]{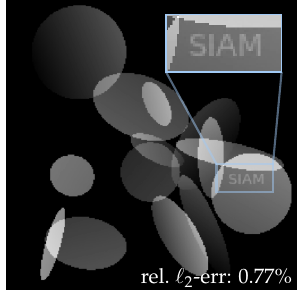}  
				&\includegraphics[width=\linewidth]{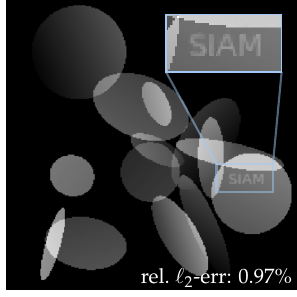}  
				&\includegraphics[width=\linewidth]{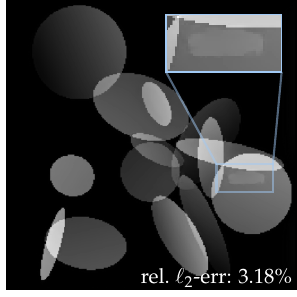} \\
			\end{tabular} \\
			\begin{tabular}{@{}>{\centering\arraybackslash}b{0.84\textwidth}@{}}   
                     {Stability towards worst-case noise} \\ 	
                
	            \setlength{\tabcolsep}{0pt} 
                \begin{tabular}{@{}>{\centering}m{0.05\textwidth}>{\centering\arraybackslash}m{0.55\textwidth}@{}} 
					\begin{turn}{90}\parbox[b]{0.28\textwidth}{\begin{center} 
                        {Relative $\ell^2$-error [\%]}
					\end{center}}\end{turn}
					&\includegraphics[width=\linewidth]{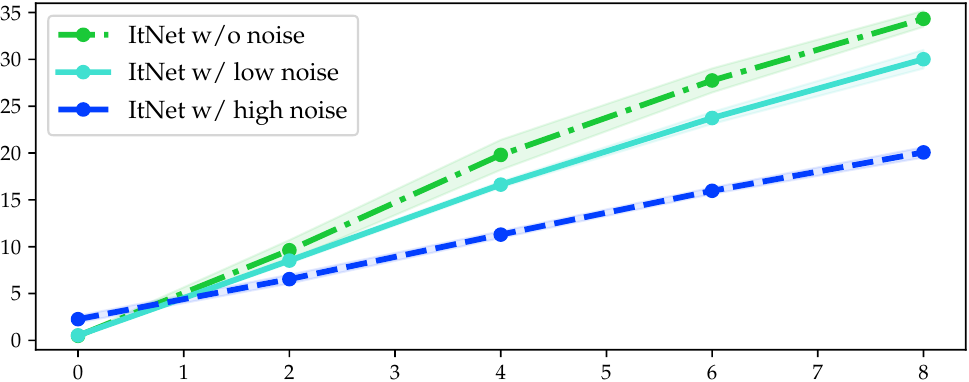} \\
                    & {Relative noise level [\%]}
				\end{tabular} \\
			\end{tabular}
		\end{tabular}
	\end{small}
	\caption{\label{fig:3}(\textbf{The accuracy-stability trade-off})
Three NNs are trained on the same dataset of images with noiseless, low-noise or high-noise measurements, respectively. The first has the highest accuracy, but the worst stability, while third has the lowest accuracy, but the best stability. The measurement matrix is a subsampled Fourier transform with $m/N \approx 17\%$.  This experiment is based on one shown in \cite{genzel2020solving}. See \S \ref{s:figures-experiments} for further information.}
\end{figure}

In Fig.\ \ref{fig:3} we show an example of this trade-off, thereby demonstrating Theorem \ref{thm:locallipcte} in practice. This figure is based on an experiment shown in \cite{genzel2020solving}.
Here, training a NN with noiseless measurements yields high performance, but high instability, while training with highly noisy measurements yields high stability, at the expense of significantly worse performance. Adding noise to the measurements before training is a simple way to balance stability and accuracy in DL. But this is not the only way to trade-off between these two competing factors. In general, developing training strategies that optimize this trade-off a challenging problem.

The second implication of \eqref{eq:cond_instability}-\eqref{eq:cond_instability2} is that $\Psi$ hallucinates. To see this, it is convenient to rewrite $x'$ as $x' = x + x_{\mathrm{Det}}$. Assume for simplicity that $x'' = x'$. Then \eqref{eq:cond_instability} states that $\Psi$ recovers both $x$ and $x + x_{\mathrm{Det}}$ well, where, due to \eqref{eq:cond_instability2}, the detail $x_{\mathrm{Det}}$ has small measurements $A x_{\mathrm{Det}}$. Part (ii) of Theorem \ref{thm:locallipcte} now asserts that $\Psi$ hallucinates. There exists a small perturbation $e$ with $\tnm{e} \leq \eta$ such that, for example, $\Psi(A x + e)$ is within $\eta$ of $x' = x + x_{\mathrm{Det}}$.  Thus, a small perturbation of the measurements of  the detail-free image $x$ causes $\Psi$ to falsely reconstruct the image $x'$ containing the detail. It consequently yields a false positive. Moreover, the result is stable, in the sense that it holds not just for $x$, $x'$ and $e$, but for any vectors lying in balls around them. Note also that it is not necessary for $\Psi$ to overperform in order for it to hallucinate. If \eqref{eq:cond_instability2} holds for some $x''$ that is not close to $x$, then (ii) implies hallucinations, in the sense that there are perturbations for which $\Psi(A x + e) \approx x''$.

The conditions of Theorem \ref{thm:locallipcte} are very general, since they pertain to the performance of $\Psi$ on two elements only. Moreover, overperformance in the above sense can easily arise when training a DNN. Indeed, if the training set contains two elements $(A x , x)$ and $(A x', x')$ for which \eqref{eq:cond_instability2} holds, then a small training error implies \eqref{eq:cond_instability}. Similarly, inconsistent performance can occur whenever $x$ belongs to the training set and $x'$ is close to this set, but not in it.

Much like with Theorems \ref{thm:AIhal2} and \ref{thm:AIhal1} (recall Remark \ref{rem:why-not-CS}), model-based methods are generally not susceptible to the conclusions of Theorem \ref{thm:locallipcte} whenever $A$ satisfies appropriate conditions. In particular, these conditions typically ensure that \eqref{eq:cond_instability} and \eqref{eq:cond_instability2} can only hold if $\nm{x - x''}_{\ell^2}$ is small, thus voiding the conclusions of the theorem. We discuss this further in \S \ref{s:model-based-CS} in the case of sparse regularization via $\ell^1$-minimization when $A$ satisfies the rNSP.

\subsection{Instabilities and hallucinations are not rare events}

Theorem \ref{thm:locallipcte} asserts the existence of `bad' perturbations, which cause either instabilities or hallucinations. It says nothing about how prevalent such bad perturbations are. However, the fact the various conclusions of Theorem \ref{thm:locallipcte} hold in balls implies that these are not rare events. We formalize this statement in the following theorem.

\begin{theorem}\label{thm:locallipcte-random}
    Let $A \in \bbC^{m \times N}$, $x,x' \in \mathbb{C}^N$ and $\eta >0$.  Let $\Psi \colon \mathbb{C}^m \rightarrow  \mathbb{C}^N$ be continuous and suppose that $\Psi$ satisfies \eqref{eq:cond_instability}--\eqref{eq:cond_instability2} for some $x'' \in \C^N$. If $E = \{E_1,\ldots, E_m\}$ is an absolutely continuous complex-valued random vector with a strictly positive probability density function, then the following hold.
\begin{itemize}[leftmargin=0.75cm]
    \item[(i)] \emph{(Instabilities are not necessarily rare events)}.
            There is a closed ball $\mathcal{B}_x\subset \C^N$ centred at $x$ and a $c > 0$ such that 
        \begin{equation}\label{eq:lower_bound}
            \mathbb{P}\big(\|\Psi(A\tilde{x}+E) - \Psi(A\tilde{x})\| \geq  \|x-x''\| - 2\eta \big) \ge c, \quad \forall \, \tilde{x} \in \mathcal{B}_x.
		\end{equation}
        Moreover, for any $0 < \delta < 1$, there is a complex Gaussian distribution on $E$ whose mean has norm at most $\eta$ such that \eqref{eq:lower_bound} holds with $c=1-\delta$.

        \item[(ii)] \emph{(Hallucinations are not necessarily rare events).} There is a $c > 0$ and $z \in \bbC^N$ with $\nm{z} \geq \nm{x-x''}$ (for example, $z = x''-x$), and closed balls $\mathcal{B}_x, \mathcal{B}_z \subset \C^N$, centred at $x$ and $z$, respectively, such that 
	\begin{equation}\label{eq:falseposprob}
		\mathbb{P}\big(\|\Psi(A \tilde{x}  +E)- (\tilde{x} + \tilde{z})\| \leq \eta\big) \ge c,\quad \forall \, \tilde{x} \in \cB_x, \tilde{z} \in \cB_{z}.
	\end{equation}
Moreover, for any $0 < \delta < 1$, there is a complex Gaussian distribution on $E$ whose mean has norm at most $\eta$ such that  \eqref{eq:falseposprob} holds with $c=1-\delta$.
	\end{itemize}
\end{theorem}
This result shows that instabilities and hallucinations are not rare events. If the perturbation is drawn randomly from an arbitrary distribution, the probability of it causing a hallucination or instability is non-zero. Furthermore, this occurs with high probability for Gaussian noise with small mean. Note that Gaussian noise is ubiquitous in imaging applications.

A limitation of this result is that it makes no claims as to the size of the variance of the Gaussian noise for which its conclusions hold. However, under somewhat more restrictive conditions one can show that similar effects occur for Gaussian noise of arbitrarily small variance as $m \rightarrow \infty$.

\begin{theorem}\label{thm:worstvar}
    Let $\nm{\cdot}$ and $\vertiii{\cdot}$ be equal to the Euclidean norm  $\nm{\cdot}_{\ell^2}$. Let $U \in \C^{N\times N}$ be unitary, $\Omega \subset \{1,\ldots, N\}$ with $m=|\Omega|$, $A = P_{\Omega} U$, $x, x'  \in \C^N$, $\eta, \delta, \sigma > 0$ and $\Psi \colon \C^m \to \C^N$ be continuous. Suppose that $\|\Psi(Ax)-x\|_{\ell^2} < \eta$, $ \|A x' - Ax\|_{\ell^2} > \delta$ and that, for every $\tilde x' \in \mathcal{B}(x', \delta)$ there is a $x'' \in \C^N$ such that 
    \begin{equation}\label{eq:bound_inst}
    \|\Psi(A\tilde x')-x''\|_{\ell^2} <  \eta.
    \end{equation}
Denote the union of all such $x''$'s by $\mathcal{S}$. Let $\xi \in \C^m$  and $I\in \C^{m\times m}$ be the identity matrix. Then, if $E \sim \C\mathcal{N}(\xi,\sigma^2 I )$ we have that 
		\begin{equation}\label{eq:lower_Gauss1}
			\begin{split}
				\mathbb{P}&\left(\|\Psi(Ax+E)-\Psi(Ax)\|_{\ell^2} \geq \inf_{x'' \in \mathcal{S}}\|x-x''\|_{\ell^2} - 2\eta\right)\\ &\geq 
			(\pi \sigma^2)^{-m} \int_{\mathcal{B}} \exp\left( -\frac{1}{\sigma^2} \left\|t - \xi\right\|_{\ell^2}^{2} \right) d t
			\end{split}
		\end{equation}
        where $\mathcal{B} \coloneqq \mathcal{B}(Ax'-Ax, \delta)$.
        Moreover, there exists a variance $\sigma_0 = \sigma_0(m)$ maximizing the lower bound in \eqref{eq:lower_Gauss1}, such that $\sigma_0(m) \rightarrow 0$ as $m, N \rightarrow \infty$ for fixed $\Omega$, $x$, $x'$, $\delta$ and $\xi$.   {Furthermore, if $\nmu{A x' - A x - \xi}_{\ell^2} < \delta$ then
\bes{
        \mathbb{P}\left(\|\Psi(Ax+E)-\Psi(Ax)\|_{\ell^2} \geq \inf_{x'' \in \mathcal{S}}\|x-x''\|_{\ell^2} - 2\eta\right) \rightarrow 1,
}
as $m \rightarrow \infty$ whenever $\sigma = \sigma(m) \leq \sigma_0(m)$.
        }
\end{theorem}

\subsection{Optimal maps may be impossible to train}

In this section, we switch focus and consider optimal recovery via the notion of \textit{optimal maps}. This study is motivated by the following question: given an inverse problem and a model class $\cM_1$, what is the best possible reconstruction map? Note that we consider noiseless measurements in this section, in order to focus the discussion on the underlying accuracy of the recovery. It is possible to extend such considerations to the noisy regime \cite{bourrier2014fundamental}.

In order to maintain sufficient generality, we consider multivalued maps in this section. Doing so allows one to consider approaches such as sparse regularization that rely on minimizers of convex optimization problems which may not be unique. Recall that a multivalued map is typically denoted with double arrows. Thus, in this section, we consider maps of the form
$
\Psi: \mathcal{M}_2 \rightrightarrows \bbC^N,
$
where $\cM_2 = A \cM_1$. We assume that the set $\Psi(y)$ is bounded for all $y \in \mathcal{M}_2$. Given a metric $d$, we use the Hausdorff distance to measure the distance between two bounded sets $X,Z \subset \mathcal{M}_1$, i.e.,
\[
d^{H}(Z,X)=\max \left \{\sup _{{x\in X}}\inf _{{z\in Z}}d(z,x),\,\sup _{{z\in Z}}\inf _{{x\in X}}d(z,x) \right \}.
\]
With slight abuse of notation we will denote a singleton $\{x\} \subset \mathcal{M}_1$ simply as $x$.

\begin{definition}[Optimal map]\label{def:opt}
Let $d$ be a metric, $\cM_1 \subset \bbC^N$, $A \in \bbC^{m \times N}$ and $\cM_2 = A \cM_1$. The optimality constant of $(A,\cM_1)$ is defined as 
	\begin{align}\label{eq:mapconst}
	c_{\mathrm{opt}}(A,\mathcal{M}_1) = \inf_{\Psi: \mathcal{M}_2 \rightrightarrows \bbC^N} \sup_{x \in \mathcal{M}_1}  d^H(\Psi(Ax),x).
	\end{align}
A map $\Psi \colon \cM_2 \rightrightarrows \bbC^N$ is an optimal map for $(A,\cM_1)$ if it attains this infimum.
\end{definition}
	
An optimal map is the best possible reconstruction map for a given sampling operator $A$ and model class $\cM_1$. However, such a map may not exist, since the infimum in \eqref{eq:mapconst} may not be attained. This motivates the following.

\begin{definition}[Approximately optimal maps]\label{def:app-opt}
Let $d$ be a metric, $\cM_1 \subset \bbC^N$, $A \in \bbC^{m \times N}$ and $\cM_2 = A \cM_1$. A family of approximately optimal maps for $(A,\cM_1)$ is a collection of maps $\Psi_{\epsilon} \colon \mathcal{M}_2 \rightrightarrows \bbC^N$, $\epsilon \in (0,1]$, such that
	\begin{equation}\label{eq:approx_optimal}
	\sup_{x \in \mathcal{M}_1}  d^H(\Psi_{\epsilon}(Ax),x) \leq c_{\mathrm{opt}}(A,\mathcal{M}_1) + \epsilon,\qquad \forall \epsilon \in (0,1].
	\end{equation}
\end{definition}

In the next result, we consider whether or not training gives rise to optimal or approximately optimal maps. Our main conclusion is that it generally does not.

\begin{theorem}
		\label{thrm:not_optimal}
		Let the metric $d$ be induced by the norm $\nm{\cdot}$, $A \in \mathbb{C}^{m \times N}$, $K \in \{2,\hdots, \infty\}$ and $\cB \subset \mathbb{C}^N$ be the closed unit ball with respect to $d$. Let $\sigma_{\min}(A) = \sigma_{\min\{m,N\}}(A)$ denote the minimum singular value of $A$ (in particular, $\sigma_{\min}(A) = 0$ if $\mathrm{rank}(A) < \min \{m,N \}$). Then the following holds.
		\begin{enumerate}[leftmargin=0.75cm,label=(\roman*)]
			\item \emph{(Training may not yield optimal maps)}. 
			There exist uncountably many $\cM_1 \subset \cB$,  such that for each $\cM_1$ there exist countably many sets $\cT \subset A \cM_1 \times \cM_1$ with $|\mathcal{T}| = K$, where $\cM_2 = A (\cM_1)$, with the following properties. Any Lipschitz continuous map $\Psi: \mathcal{M}_2 \rightarrow \bbC^N$  (potentially multivalued $\Psi: \mathcal{M}_2 \rightrightarrows \bbC^N$) with Lipschitz constant $L$ that satisfies $\sigma_{\min}(A)L <1$ and for which 
			\begin{equation}\label{eq:the_goal1}
				d^H(\Psi(y) ,x) \leq \delta, \qquad \forall (y,x) \in \mathcal{T},
			\end{equation}
for some $\delta < \frac{1-L\sigma_{\min}(A)}{4}$, is not an optimal map. Moreover, the collection of such mappings does not contain a family of approximate optimal maps. If $K$ is finite, one can choose $|\mathcal{M}_1| = K+1$, in which case there is at least one $\cT$ with the above property.
			\item \emph{(The map sought by training may not exist)}. There exist uncountably many $\mathcal{M}_1 \subset \cB$ with $|\mathcal{M}_1| = K$ such that, for $\cM_2 = A \cM_1$, there does not exist 
			a Lipschitz continuous map $\Psi: \mathcal{M}_2 \rightarrow \bbC^N$ (nor a multivalued map $\Psi: \mathcal{M}_2 \rightrightarrows \bbC^N$) for which $\sigma_{\min}(A)L <1$, $\delta < \frac{1-L\sigma_r(A)}{4}$ and
			\begin{equation*}
				d^H(\Psi(y) ,x) \leq \delta, \qquad \forall (y= Ax,x) \in \cM_2 \times \mathcal{M}_1.
			\end{equation*}
		\end{enumerate}
\end{theorem}
A well-trained reconstruction map $\Psi$ should satisfy \eqref{eq:the_goal1} for a suitable $\delta$ and some suitable collection of images $\cT$ (e.g., the training or test set, or a subset thereof). Thus, part (i) of this theorem states that successful training may not yield an optimal map or an approximately optimal map. Part (ii) shows that it may not be possible to achieve a small error over of $\cM_1$ in the first place -- i.e., there are model classes that cannot be (implicitly) learned. We remark in passing that the (mild) conditions $\delta \leq \frac{1-L\sigma_{\min}(A)}{4}$ and $\sigma_{\min}(A)L <1$ are related to the assumption that $\cM_1$ is a subset of the unit ball $\cB$. The theorem holds larger $\delta$, provided this ball is suitably enlarged.

Much like Theorem \ref{thm:AIhal1}, Theorem \ref{thrm:not_optimal} is not simply a statement about overfitting (Remark \ref{not-overfitting} also applies in this case). In particular, Theorem \ref{thrm:not_optimal} implies that $\Psi$ can have excellent performance on both the training and test sets but still be suboptimal.

\begin{remark}
To understand Theorem \ref{thrm:not_optimal} better, it is worth contrasting it with the case of classification. Consider a binary classification problem with an unknown ground truth labelling function $f \colon \cM \rightarrow \{ \pm 1 \}$. Then the training data
$
\{ (x_i , f(x_i) ) \}^{K}_{i=1}
$
 consists of a finite subset of the graph $\{ (x , f(x) ) : x \in \cM \}$
of this function. Thus, in classification, we learn a NN approximation to $f$ from a finite sample of its graph.
By contrast, in an inverse problem the training set has the form
\be{
\label{trainingdatasec5}
\{ (A x_i , x_i) \}^{K}_{i=1}.
} 
However, part (ii) of Theorem \ref{thrm:not_optimal} implies that this generally does not correspond to a finite subset of the graph $\{ (y,\Phi(y)) : y \in \cM_2 \}$ of some optimal (or approximately optimal) map $\Phi$. Thus, there is no reason why training should yield an optimal or approximately optimal map, since the training data \eqref{trainingdatasec5} is not sampled from the graph of such a map. Indeed, this is exactly what is asserted in part (i) of the theorem.

Why does this scenario arise? The answer lies in the nature of $A$.  Suppose that $A \in \bbC^{m \times N}$ has $\mathrm{rank}(A) = N$ and is well conditioned. Then an optimal map with small Lipschitz constant is simply $f(y) = A^{\dag} y$, where $A^{\dag}$ denotes the pseudoinverse of $A$, and the training data \eqref{trainingdatasec5} is then a subset of the graph of $f$.
 Unfortunately, $A$ is either ill-conditioned or has $\mathrm{rank}(A) < N$ in practical imaging scenarios.
\end{remark}

\section{Proofs of the main results}\label{s:dismain}
We now prove the main results.

\subsection{Proofs of Theorem \ref{thm:AIhal2} and Theorem \ref{thm:AIhal1}}

\begin{proof}[Proof of Theorem \ref{thm:AIhal2}]
	For part (i), let $ e \in \mathcal{B}(0, \delta)$. Then
	\begin{equation*}\label{eq:appx1}
		\Psi(Ax+ e) = \Psi(A(x+x_{\mathrm{Det}})+(e- Ax_{\mathrm{Det}})) = \Psi(A(x+x_{\mathrm{Det}}))+z',
	\end{equation*}
	where 
$
			\|z'\| = \|\Psi(A(x+x_{\mathrm{Det}})+(e- Ax_{\mathrm{Det}})) - \Psi(A(x+x_{\mathrm{Det}}))\|
            \leq L \vertiii{e- Ax_{\mathrm{Det}}} \leq 2L\delta.
$
	Furthermore, by assumption we have that 
	$\Psi(A(x+x_{\mathrm{Det}})) = x+x_{\mathrm{Det}} + \tilde{z}$,
	where
		$\|\tilde{z}\| = \|\Psi(A(x+x_{\mathrm{Det}})) - (x+x_{\mathrm{Det}})\| \leq \delta$.
Combining, we get that 
$$
\Psi(Ax+e) = x+x_{\mathrm{Det}}+ z_e,
\quad	\text{where }\|z_{e}\| = \| z'+\tilde{z}\| \leq (1+2L)\delta.
$$
Next consider part (ii). Given the family $\cN\cN$ as in \eqref{NN-family}, let $\Phi \in \cF$ be arbitrary, $W \in \bbC^{N \times n}$ be the zero matrix and $b = x + x_{\mathrm{Det}}$. Then let $\widetilde{\Psi} : y \mapsto W \Phi(y) + b$. Notice that $\widetilde{\Psi} \in \cN \cN$ and $\widetilde{\Psi}(y) = x + x_{\mathrm{Det}}$, $\forall y \in \bbC^m$. Hence $\widetilde{\Psi}$ is constant, and therefore Lipschitz continuous with constant $0 < L$, and satisfies \eqref{eq:NN_exists}.
\end{proof}

\begin{proof}[Proof of Theorem \ref{thm:AIhal1}]
Let $r =\mathrm{rank}(A)$ and denote the singular values of $A$ as $\sigma_{1}(A) \geq \cdots \geq  \sigma_{\min\{m,N\}}(A) = 0$, where $\sigma_i(A) = 0$ for $r < i \leq \min \{m,N \}$ whenever $r < \min\{m,N\}$. Since all norms on finite dimensional vector spaces are equivalent, let $K_1,K_2 > 0$ be such that 
    \begin{equation} \label{eq:norm_ineq333}
       \|u\|_{\ell^2} \leq K_1 \vertiii{u},\quad \forall u \in \C^m, \quad\text{and}\quad 
       \|v\| \leq K_2\|v\|_{\ell^2}, \quad \forall v \in \C^N.
    \end{equation}    
    We start by constructing $\mathcal{M}_1$.  Since $\mathrm{rank}(A) \geq 1$, we can find a vector 
\be{
\label{eta-def}
\eta \in
    	\begin{cases}
    		 \mathcal{N}(A)^{\perp} \qquad &r< \min\{m,N\} \\
    		 \mathbb{C}^N    \qquad    &r= \min\{m,N\}
    	\end{cases}
}
such that $A(\eta + x_{\mathrm{Det}}) \neq 0$ and $\vertiii{A\eta} = \widetilde \epsilon$, where $\widetilde \epsilon < \min\{\delta/(4L), \sigma_r(A) \delta/(2K_1 K_2)\}$. 
    Fix $x'\in \mathcal{T}$ and let 
    \begin{equation}\label{eq:defM_1}
\mathcal{M}_1 = \mathcal{T} \cup \{x' +e^{n \mathrm{i}}(\eta+x_{\mathrm{Det}})\}_{n=1}^{\infty}.
	\end{equation}
    Observe that $A\mathcal{M}_1 \subset (A\mathcal{T})^{\epsilon}$, as $\mathcal{T}\subseteq \mathcal{M}_1$ and $\vertiii{Ax'- A(x' +e^{n \mathrm{i}}(\eta+x_{\mathrm{Det}}))} = \vertiii{A(\eta + x_{\mathrm{Det}})} < \delta/(2L) \leq \epsilon$ for any $n \in \mathbb{N}$. To get a set $\mathcal{M}_1$ with finite cardinality simply choose an appropriate subset in \eqref{eq:defM_1}. To create the uncountable family $\mathcal{C}$, pick any other value for $\widetilde{\epsilon} \in (0,\min\{\delta/(4L), \sigma_{r}(A)\delta/(2K_1K_2))\})$ and repeat the construction. 
	
	Next let $n\in \mathbb{N}$ and notice that 
    \begin{equation*}\label{eq:bound432}
        \begin{split}
             \| \Psi(A(x' + &e^{n\mathrm{i}}(\eta + x_{\mathrm{Det}}))) - x'\| \\
            \leq&~ \|\Psi(A(x'+e^{n\mathrm{i}}(\eta+ x_{\mathrm{Det}}))) - \Psi(Ax')\| + \|\Psi(Ax') - x'\| \\
            \leq&~ L \vertiii{A(x'+e^{n\mathrm{i}}(\eta +x_{\mathrm{Det}}))- Ax'} + \delta \leq \tfrac{3}{2}\delta.
        \end{split}
    \end{equation*}
Next we claim that for
\begin{align*}
\xi \in
	\begin{cases}
		\mathcal{N}(A)^{\perp} \qquad &r< \min\{m,N\} \\
		\mathbb{C}^N    \qquad    &r= \min\{m,N\}
	\end{cases}
\end{align*}
we have $\|\xi\|_{\ell^2} \leq \nm{A \xi}_{\ell^2}/\sigma_r(A)$. To see this, let $u_i \in \C^{m}$ and $v_i \in \C^{N}$, $i=1,\ldots, \min\{m,N\} $ denote the left and right singular vectors of $A$, respectively. Write $\xi = \lambda_1 v_1 +\cdots +\lambda_r v_r$, for scalars $\lambda_i$ and notice that $A\xi = \sigma_1 \lambda_1 u_1 + \cdots  + \sigma_r \lambda_r u_r$, where $\sigma_i = \sigma_i (A)$. Since the $u_i$'s and $v_i's$ are orthonormal, we have $\|A\xi\|_{\ell^2}^{2} = \sigma_1^{2}\lambda_1^2 + \ldots + \sigma_r^{2}\lambda_r^{2}$ and $\|\xi\|_{\ell^2}^2 = \lambda_{1}^{2} +\cdots+\lambda_{r}^{2}$. Since $\sigma_i/\sigma_r \geq 1$ for $i = 1,\ldots,r$ the inequality follows.

Now, since $\vertiii{A\eta} = \tilde \epsilon$, we know from \eqref{eq:norm_ineq333} that $\|A\eta\|_{\ell^2}\leq K_1 \vertiii{A\eta} = K_1 \tilde{\epsilon} < \sigma_r(A)\delta/(2K_2)$. By \eqref{eta-def}, the above claim and \eqref{eq:norm_ineq333} we get $\|\eta\| \leq K_2\|\eta\|_{\ell^2} \leq \delta/2$. 

Next let $\lambda = -e^{n\mathrm{i}}$. Then 
    \begin{equation}\label{eq:bound932}
        \begin{split}
        &\|\Psi(A(x' + e^{n\mathrm{i}}(\eta+x_{\mathrm{Det}}))) 
            - (x' + e^{n\mathrm{i}}(\eta+x_{\mathrm{Det}}) 
            + \lambda x_{\mathrm{Det}})\| \\ 
       & \leq \|\Psi(A(x' + e^{n\mathrm{i}}(\eta+x_{\mathrm{Det}}))) - x'\| + \|e^{n\mathrm{i}}\eta\| \leq 2\delta. 
        \end{split}
	\end{equation}
By assumption $\mathbb{P}_{X\sim \mathcal{D}}(X\in \mathcal{M}_1\setminus \mathcal{T}) \geq 1-q$. Now, since any $X \in \mathcal{M}_1\setminus \mathcal{T}$ is of the form $X=x'+e^{n\mathrm{i}}(\eta+x_{\mathrm{Det}})$ part (i) of the result now follows from \eqref{eq:bound932}.  
  	
Next we show (ii) for our choice of $\mathcal{M}_1 \in \mathcal{C}$. Recall that by \enquote{algorithm} we mean a Blum-Shub-Smale (BSS) machine \cite{BCSS}. In particular, this means that the algorithm can perform arithmetic operations with real numbers, and  check if two real numbers are equal. We describe the algorithm $\Gamma(y)$, taking an input $y \in A\mathcal{M}_1$ and yielding a NN $\Phi_{y} \colon \C^{m} \to \C^N$ as the output of the following pseudo-code:

\renewcommand{\algorithmicend}{end}
\renewcommand{\algorithmicif}{if}
\renewcommand{\algorithmicthen}{then}
\renewcommand{\algorithmicelse}{else}
\renewcommand{\algorithmicfor}{for}
\renewcommand{\algorithmicforall}{for all}
\renewcommand{\algorithmicdo}{do}
\renewcommand{\algorithmicwhile}{while}
\renewcommand{\algorithmicreturn}{return}
\algsetup{indent=2.5em}
\begin{algorithmic}[1]
    \FORALL{$x\in \mathcal{T}$ }
    \IF{$y = Ax$}
        \STATE{$\Phi_y \gets \Psi$ and return $\Phi_y$;}
    \ENDIF
    \ENDFOR
    \STATE{set $l=1$ and $u = A(\eta+x_{\mathrm{Det}})$;}
    \WHILE{$u[l] = 0$}
        \STATE{$l\gets l + 1$;}
    \ENDWHILE
    \STATE{set $c = (y[l] - (Ax')[l])/u[l]$;}
    \STATE{let $\Phi_y$ be a such that $\Phi_y(y) = x'+c(\eta +x_{\mathrm{Det}})$ and return $\Phi_y$. }        
\end{algorithmic}
    A few comments are in order. The set $\mathcal{T}$ is finite, and the first for loop will, therefore, either terminate or return the desired NN. If not, then the while loop finds an index $l \in \{1,\ldots,m\}$ such that $A(\eta+x_{\mathrm{Det}})[l] = u[l] \neq 0$. Such an index exists since $A (\eta+x_{\mathrm{Det}}) \neq 0$ by construction. Next, notice that if $x \in \cM_1 \backslash \cT$, then $x = x' + e^{\mathrm{in}}(\eta + x_{\mathrm{Det}})$ for some $n\in \mathbb{N}$. Hence, if $y=A x$, then 
    \[
    c =   \frac{(y - Ax')[l]}{u[l]}
      = \frac{(A(x' + e^{\mathrm{i}n}(\eta + x_{\mathrm{Det}})) - Ax')[l]}{(A(\eta  + x_{\mathrm{Det}}))[l]}
      = e^{\mathrm{i}n} .
    \]
 Finally, in the case $x \notin \cT$ we choose
 $\Phi_{y} \colon \C^{m} \to \C^N$ from $\cN\cN$ as in \eqref{NN-family} as follows. Let $\tilde{\Phi} \in \cF$ be arbitrary, $W \in \bbC^{N \times n}$ be the zero matrix and $b= x' + c(\eta+x_{\mathrm{Det}})$.
Since $c=e^{\mathrm{i}n}$, this means that $\Phi_y(y) = x' + c(\eta+x_{\mathrm{Det}})$ will identify the vector $x$ whenever $x \in \cM_1\backslash \cT$.
\end{proof}

\subsection{Proofs of Theorems \ref{thm:locallipcte}, \ref{thm:locallipcte-random} and \ref{thm:worstvar}}

\begin{proof}[Proof of Theorem \ref{thm:locallipcte}]
    Consider part (i). Let $y=Ax$ and note that $\tnm{Ax - A x' } \leq \eta \leq \epsilon$. We apply this fact, together with the reverse triangle inequality and the assumption \eqref{eq:cond_instability} to get that  
    \begin{equation}\label{eq:lipbound}
    \begin{split}
        L^{\eps}(\Psi, y) &= \sup_{\substack{e \in \C^m \\ \vertiii{e} \leq \epsilon}} \frac{\|\Psi(y) - \Psi(y+e)\|}{\vertiii{e}} 
\geq \frac{1}{\eta} \nm{\Psi(Ax) - \Psi(Ax') } \\ 
    &\geq \frac{1}{\eta} (  \nm{x'' - x} - \nm{\Psi(Ax)- x} - \nm{\Psi(Ax') - x''})
    >\frac{1}{\eta} (\|x''-x\| - 2\eta).
    \end{split}
\end{equation}
    The continuity of $\Psi$ and the strict inequality in \eqref{eq:lipbound} now implies that there exists a closed non-empty ball $\mathcal{B} \subset \mathbb{C}^m$ centred at $y = Ax$ such that, 
    $L^{\eps}(\Psi,\tilde y) \geq \tfrac{1}{\eta}(\|x-x''\| - 2\eta )$
for all $\tilde y \in \mathcal{B}$. This gives the result.

For part (ii), let $z = x''-x$ and $e = Ax'-Ax$. Clearly, $\|z\|\geq \|x-x''\|$ and $\vertiii{e}\leq \eta$. Moreover, we have that
 $   \|\Psi(A x + e) - (x+z)\| = \|\Psi(A x') - x''\| < \eta$. 
The result once more follows from continuity of $\Psi$. 
\end{proof}

\begin{proof}[Proof of Theorem \ref{thm:locallipcte-random}]
Consider part (i). Let $e = Ax' - Ax$ and notice by the triangle inequality and \eqref{eq:cond_instability} that $\|\Psi(Ax+e) - \Psi(Ax)\| > \|x-x''\|-2\eta$.
The continuity of $\Psi$ implies that there is a closed ball $\mathcal{B}_x \subset \C^N$ centered at $x$ and an $\epsilon > 0$ such that
    \[ \|\Psi(A\tilde{x}+\tilde e) - \Psi(A\tilde{x})\| \geq \|x-x''\|-2\eta \quad \forall \, \tilde{e} \in \mathcal{B}(e,\epsilon), \tilde{x}\in \mathcal{B}_x. \]
    Now, by assumption $E$ has a probability density function $g \colon \C^m \to \R_{>0}$ which is strictly positive. Thus, if $\mu$ denotes the Lebesgue measure,
    \begin{align*}
        \mathbb{P}\big(\|\Psi(A\tilde x+E) - \Psi(A\tilde x)\| \geq  \|x-x''\| - 2\eta \big) \geq \mathbb{P}(E \in \mathcal{B}(e,\epsilon)) = \int_{B(e,\epsilon)} g \mathrm{d}\mu > 0, 
    \end{align*}
for all $\tilde x \in \mathcal{B}_x$, as required.

    For the second part of the statement note that $\vertiii{e} \leq \eta$ by \eqref{eq:cond_instability2}. Furthermore, let $I \in \C^{m\times m}$ denote the identity matrix, $\sigma > 0$ and let $E \sim \C\mathcal{N}(e, \sigma^2 I)$. Then 
\begin{equation}
	\label{eq:gauss1}
        \begin{split}
\mathbb{P}(E \in \mathcal{B}(e,\epsilon)) &= \frac{1}{(\pi\sigma^2 )^{m}}\int_{\mathcal{B}(e,\epsilon)} \exp\left(-\frac{1}{\sigma^2 }\|t-e\|_{\ell^2}^{2}\right) dt \\ &= \frac{1}{\pi^m} \int_{\mathcal{B}(0, \epsilon/\sigma)} \exp\left( - \|u\|_{\ell^2}^{2}\right) du, 
        \end{split}
\end{equation}
where in the second equality we applied the change of variables $t = \sigma u + e$.
    From \eqref{eq:gauss1}, we observe that $\mathbb{P}(E \in \mathcal{B}(e,\epsilon)) \to 1$ as $\sigma \to 0$. Thus, for any choice of $\delta \in (0,1)$ we can find a $\sigma$ such that \eqref{eq:lower_bound} holds with $c = 1-\delta$, as required.

    Part (ii) is proved analogously to {part (ii)} from Theorem \ref{thm:locallipcte}, {combined with the arguments above}.
\end{proof}

\begin{proof}[Proof of Theorem \ref{thm:worstvar}]
First observe that $AA^* = P_{\Omega}UU^*P_{\Omega}^* = I$, where $I$ is the identity matrix, since $U$ is unitary and $P_{\Omega} \in \C^{m\times N}$ is a projection. It follows that $\|Au\|_{\ell^2}\leq\|u\|_{\ell^2}$ for all $u\in \C^N$ and $\|A^*v\|_{\ell^2} = \|v\|_{\ell^2}$ for all $v \in \C^m$. In particular, $A\tilde{x}'-Ax \in \mathcal{B}(Ax'-Ax, \delta)$ if $\tilde x' \in \mathcal{B}(x',\delta)$. Moreover, the converse is also true. That is, if $\tilde e \in \mathcal{B}({Ax'-Ax}, \delta)$ then there exists a $\tilde x' \in \mathcal{B}(x',\delta)$ such that $A\tilde x'-Ax = \tilde e$. Indeed, by assumption $\tilde e = A x'-Ax+v$, where $\|v\|_{\ell^2} \leq \delta$, but then we have that ${\tilde{x}' = x' + A^*v }\in \mathcal{B}( x', \delta)$.

This means that for every $\tilde{e} \in \mathcal{B}({Ax'-Ax}, \delta)$ we can find a corresponding $\tilde{x} \in \mathcal{B}(x',\delta)$ such that
    \begin{align*}
        \|\Psi(Ax+\tilde e)-\Psi(Ax)\|_{\ell^2} &=  \|\Psi(A \tilde x') -\Psi(Ax)\|_{\ell^2} > \|x - x''\|_{\ell^2} - 2\eta,
    \end{align*}
    where the last inequality follows from the assumption $\|A\tilde x ' - x''\|_{\ell^2} < \eta$. Moreover, since for every $\tilde{x} \in \mathcal{B}(x',\delta)$ we can find a $x''$ satisfying the above inequality, it is clear that by taking the infimum over all possible such $x''$ we obtain a lower bound for any choice of $\tilde{e} \in \mathcal{B}(Ax'-Ax,\delta)$, i.e., 
\begin{equation}\label{eq:prob_lb}
    \|\Psi(Ax+\tilde e)-\Psi(Ax)\|_{\ell^2} \geq \inf_{x'' \in \mathcal{S}}\|x-x''\|_{\ell^2} - 2\eta \quad \text{for all }\tilde e \in \mathcal{B}(Ax'-Ax, \delta).
\end{equation}
Next consider  $E \sim \C\mathcal{N}(\xi,\sigma^2 I)$ and observe that  
	\begin{equation}\label{eq:lower_bound_B}
		\begin{split}
			\mathbb{P}\bigg( & \|\Psi(Ax+E)-\Psi(Ax)\|_{\ell^2} \geq \inf_{x'' \in \mathcal{S}}\|x-x''\|_{\ell^2} - 2\eta\bigg) \\
			& \geq \mathbb{P}\left(E \in \mathcal{B}(A x'-Ax, \delta)\right) 
			= (\pi \sigma^2)^{-m} \int_{\mathcal{B}} \exp\left( -\frac{1}{\sigma^2} \left\| t - \xi\right\|_{\ell^2}^{2} \right) dt.
		\end{split}
	\end{equation}
This establishes \eqref{eq:lower_Gauss1}. 

    We proceed by considering the claim of the existence of a variance $\sigma_0$ maximizing \eqref{eq:lower_Gauss1}.  
Let $\mu(\mathcal{B})$ denote the Lebesgue measure of $\mathcal{B}$ and let $\gamma = Ax' - Ax - \xi$. Now observe that for $t \in \mathcal{B} = \mathcal{B}(Ax'-Ax, \delta)$ we have
    \begin{equation}\label{eq:gamma_ineq}
   \|\gamma\|_{\ell^2} - \delta \leq \|t -\xi\|_{\ell^2} \leq \|\gamma\|_{\ell^2} +\delta. 
\end{equation}
    We claim that the mapping 
	\begin{equation}\label{eq:maximum}
        \sigma \mapsto (\pi \sigma^2)^{-m}\int_{\mathcal{B}} \exp\left(-\frac{1}{\sigma^2}\|t-\xi\|^{2}_{\ell^2} \right) dt
	\end{equation}
    has a maximum over $\sigma > 0$. To see this, observe that 
    \begin{equation} \label{eq:up_bd_sig}
        (\pi \sigma^2)^{-m}\int_{\mathcal{B}} \exp\left(-\frac{1}{\sigma^2}\|t -\xi\|^{2}_{\ell^2} \right) dt \leq (\pi\sigma^2)^{-m} \mu(\mathcal{B})\exp\left(-\sigma^{-2}(\|\gamma\|_{\ell^2} - \delta)^2\right)
    \end{equation}
    and notice that the upper bound in \eqref{eq:up_bd_sig} tends to zero when $\sigma \to 0$ and when $\sigma \to \infty$. This means that we can restrict the domain of the mapping in \eqref{eq:maximum} to a closed interval when searching for the maximum. The existence of the maximum then follows by the continuity of \eqref{eq:maximum} and the Extreme Value Theorem. 

    Next, observe that the derivative of \eqref{eq:maximum} is given by
    \begin{equation}\label{eq:diff_intergral}
        \begin{split}
            &\frac{d}{d\sigma}\left[
        (\pi \sigma^2)^{-m} \int_{\mathcal{B}} \exp\left( -\frac{1}{\sigma^2} \left\|t - \xi\right\|_{\ell^2}^{2} \right) d t 
        \right] \\
        =&
            \frac{1}{\pi^m\sigma^{2m+1}}\int_{\mathcal{B}}\exp\left(-\frac{1}{\sigma^2}\|t-\xi\|^{2}_{\ell^2}\right) \left(\frac{2\|t-\xi\|^{2}_{\ell^2}}{\sigma^2} -2m \right) d t
        \end{split}
	\end{equation}
    Now, by setting \eqref{eq:diff_intergral} equal to zero, and rearranging the terms we get an expression for the $\sigma_{0}$ maximizing \eqref{eq:maximum},
    \begin{equation}\label{eq:sigma_eq}
        \sigma_{0}^2 = \frac{
    \int_{\mathcal{B}}\exp\left(-\frac{1}{\sigma^2_0}\|t-\xi\|^{2}_{\ell^2}\right) \|t-\xi\|^{2}_{\ell^2}d t
    }{m\int_{\mathcal{B}}\exp\left(-\frac{1}{\sigma^2_0}\|t-\xi\|^{2}_{\ell^2}\right) d t
    }.
\end{equation}
Using \eqref{eq:gamma_ineq} we see that
    \begin{equation}
        \sigma^2_0 \leq \frac{\mu(\mathcal{B})\exp\left(-\sigma_{0}^{-2}(\|\gamma\|_{\ell^2} - \delta)^2\right)\left(\|\gamma\|_{\ell^2} + \delta\right)^2}{m \mu(\mathcal{B})\exp\left(-\sigma_{0}^{-2}(\|\gamma\|_{\ell^2} + \delta)^2\right)} = \frac{\exp\left(4\sigma_{0}^{-2} \|\gamma\|_{\ell^2}\delta\right) (\|\gamma\|_{\ell^2}+\delta)^2}{m}.
    \end{equation}
Rearranging the terms yields
    \[
        0 \leq \sigma_{0}^2\exp(-4\sigma_{0}^{-2}\|\gamma\|_{\ell^2}\delta) \leq (\|\gamma\|_{\ell^2}+\delta)^2 / m,
    \]
which implies that $\sigma_0 \to 0$ when $m \to \infty$. 

{
By assumption $\nm{Ax' - A x - \xi}_{\ell^2} < \delta$. Hence, we must have $\cB(\xi,\delta') \subseteq \cB$ for some $\delta' \leq \delta$.
Next, apply the change of variables $t = \sigma u + \xi$ so that the integral in \eqref{eq:lower_Gauss1} can be bounded below by
\bes{
\pi^{-m} \int_{\cB'} \exp \left ( -\nm{u}^2_{\ell^2} \right ) d u,\qquad \text{where $\cB' = \cB(0,\delta'/\sigma)$}.
}
Recall that $\pi^{-m} \int_{\bbC^m} \exp(-\nm{u}^2_{\ell^2}) d u = 1$. Hence, for each $m$, let $\sigma = \sigma(m)$ be sufficiently small so that this integral is $\geq 1-1/m$. The result now follows.
}
\end{proof}

\subsection{Proof of Theorem \ref{thrm:not_optimal}}

\begin{proof}[Proof of Theorem \ref{thrm:not_optimal}]
	We begin with the proof of (ii).  Let $\{x_1, \hdots, x_{K} \}$ be $K$ distinct elements in $\cN(A)^{\perp}$ such that 
	$\|x_1\| = 1/2$ and $0 < \|x_j\| \leq 1$. Note that we can do this since $r =\mathrm{rank}(A) \geq 1$. Let $\sigma_{1}(A) \geq \cdots \geq \sigma_r(A)> 0$ be the first $r$ singular values of $A$ and $u_r \in \C^{m}$ and $v_r \in \C^{N}$ be left and right singular vectors, respectively, corresponding to $\sigma_r(A)$. Now define
\begin{align}
			z_1 =
			\begin{cases}
				v \in \cN(A) & \qquad \text{if} \qquad r < \min\{m,N\}\\
				1/2 v_r & \qquad \text{if} \qquad r = \min\{m,N\},
			\end{cases}
		\end{align}
		where $\|z_1\|=1/2$.  Note that for $r = \min\{m,N\}$ we have that  $\|Az_1\| = 1/2 \sigma_r(A)$. Let $\mathcal{M}_1 = \{x_1 + z_1,x_1,\hdots, x_{K}\}$ and observe that $\mathcal{M}_1 \subset \mathcal{B}$.  We argue by contradiction and suppose that there exists a (possibly multivalued) map $\Psi: \mathcal{M}_2 \rightrightarrows \C^N$ with 
	\begin{equation}\label{eq:Phi_on_Ax1}
		d^H(\Psi(Ax),x) \leq \delta, \qquad \forall x \in \mathcal{M}_1.
	\end{equation}
In particular, $d^H(\Psi(A x_1), x_1) \leq \delta$.  However, we have 
		$\Psi(A(x_1 + z_1)) = \Psi(Ax_1+Az_1) = \Psi(Ax_1)+\xi$ with 
		\begin{align}
			\|\xi\| \leq \begin{cases}
				0 & \qquad \text{if} \qquad r < \min\{m,N\}\\
				L/2 \sigma_r(A) & \qquad \text{if} \qquad r = \min\{m,N\},
			\end{cases}
		\end{align}
		by the Lipschitz continuity of $\Psi$. Therefore by translation invariance of the metric, we get $d^H(\Psi(A(x_1 + z_1)), x_1 + z_1) = d^H(\Psi(Ax_1), x_1 + z_1-\xi) \geq \nm{z_1} -L/2\sigma_r(A)- \delta \geq 1/2 -L/2\sigma_r(A) - \delta > \delta $, which contradicts \eqref{eq:Phi_on_Ax1}.
	
	Since $K$ is arbitrary, in both of the above cases we get the result.  In order to get uncountably many different $\mathcal{M}_1$'s, as mentioned in the statement of the theorem, one can simply multiply the original choice of $\mathcal{M}_1$ by complex numbers of modulus $1$. 
	
To prove (i) we use the setup from the proof of (ii). Indeed, let $\mathcal{M}_1$ be as defined previously, and set $\mathcal{T} = \{x_1,\hdots, x_K\}$.  First, we consider the case where $r =\mathrm{rank}(A)$ with $r < \min\{m,N\}$.
		Define the map $\psi_0: \mathcal{M}_2 \rightarrow \C^N$ by
		\begin{equation}\label{eq:cases_psi}
			\psi_0(y) =
			\begin{cases}
				x_1 + \frac{1}{2}z_1& \text{if } y = Ax_1 \\
				x_1 + \frac{1}{2}z_1 & \text{if } y = A(x_1+z_1) \\
				x_j & \text{otherwise} \\
			\end{cases} .
		\end{equation}
		Then, by \eqref{eq:cases_psi},
		\begin{equation}\label{eq:optimal_bound1}
			\begin{split}
				c_{\mathrm{opt}}(A,\mathcal{M}_1) &= \inf_{\varphi: \mathcal{M}_2 \rightrightarrows \mathcal{M}_1} \sup_{x \in \mathcal{M}_1}  d^H_1(\varphi(Ax),x) \leq  \sup_{x \in \mathcal{M}_1}  \|\psi_0(Ax)-x\| \\
				& \quad \leq \sup_{j \geq 1}  \|\psi_0(Ax_j)-x_j\| \vee \|\psi_0(A(x_1+z_1))-(x_1+z_1)\| = \frac{1}{4},
			\end{split}
	\end{equation}
where $a \vee b$ denotes the standard maximum of real numbers $a,b$.  
	However, for any mapping $\Psi: \mathcal{M}_2 \rightrightarrows \mathbb{C}^N$ with 
	\begin{equation}
		\label{greenhouse}
		d^H(\Psi(A x_j), x_j) \leq \delta,\quad \forall j = 1,\ldots,K,
	\end{equation}
	we have that 
	\begin{equation*}
		\begin{split}
			& \sup_{x \in \mathcal{M}_1}  d^H(\Psi(Ax),x)  \geq d^H(\Psi(A(x_1 +z_1)),x_1+z_1) \\
			& \qquad = 
			d^H(\Psi(Ax_1 ), x_1+z_1) \geq  \|z_1\| - d^H(\Psi(Ax_1 ),x_1) \geq 1/2 - \delta \\
			&  \qquad > 1/4.
		\end{split}
	\end{equation*}
	Thus, by \eqref{eq:optimal_bound1}, it follows that $\Psi$ is not an optimal map.  Furthermore, it is clear that no family of maps satisfying \eqref{greenhouse} can be approximately optimal.
	
Now we consider the case $r =\mathrm{rank}(A)$ with $r = \min\{m,N\}$. Define the map $\psi_0: \mathcal{M}_2 \rightarrow \C^N$ by
		\begin{equation}\label{eq:cases_psi0}
			\psi_1(y) =
			\begin{cases}
				x_1 & \text{if } y = Ax_1 \\
				x_1 + z_1 & \text{if } y = A(x_1+z_1) \\
				x_j & \text{otherwise} \\
			\end{cases} .
		\end{equation}
		Then, we have that $c_{\mathrm{opt}}(A,\mathcal{M}_1) = 0$.  However, for any mapping $\Psi: \mathcal{M}_2 \rightrightarrows \mathbb{C}^N$  satisfying \eqref{greenhouse}, as before we have that 
		\begin{equation*}
			\begin{split}
				& \sup_{x \in \mathcal{M}_1}  d^H(\Psi(Ax),x)  \geq d^H(\Psi(A(x_1 +z_1)),x_1+z_1) \\
				&  = d^H(\Psi(Ax_1), x_1 + z_1-\xi) >\delta > 0.
			\end{split}
		\end{equation*} 
		Thus, it follows that $\Psi$ is not an optimal map.
\end{proof}

\section{Conclusions and prospects}

This paper has strived to provide theoretical explanations for the growing concerns surrounding the use of AI-based methods for inverse problems in imaging. Our main results describe mathematical mechanisms that cause reconstruction maps to hallucinate, become unstable or, in a general sense, perform in an unpredictable or inconsistent manner. While the motivations for this work were AI-based methods, we reiterate at this stage that many of our results are general, and therefore apply more broadly (recall Remark \ref{rem:sparse-reg-also}). Nevertheless, our findings indicate that learning-based methods may be especially susceptible to such phenomena. For example, training can easily lead to \textit{overperformance}, yielding instabilities and hallucinations. How to best balance this \textit{accuracy-stability trade-off} while learning is a key problem for future investigations. We remark also that our results apply to discrete linear inverse problems that are ill-conditioned or ill-posed in the sense that the solution is nonunique. It is interesting to investigate how other sources of ill-posedness such as discontinuity of the forward map may also lead to such effects in learning-based methods.

We end this paper with a discussion of the broader context and future prospects.

\subsection{Adversarial perturbations and instabilities in AI}

When adversarial attacks and instabilities were first discovered in late 2013 for image classification \cite{SzZ-14}, the general sentiment was that this issue would be quickly solved. This sentiment is maybe best exemplified by Turing Award laureate Geoffrey Hinton's famous quote \enquote{\textit{They should stop training radiologists now.}} (The New Yorker, 2017) \cite{Hinton_quote}. Yet, what happened in the aftermath this discovery was an arms race between those developing new defence strategies and those developing new attack algorithms \cite{ortiz2021optimism}. While this has spurred many new developments and insights into the robustness of DL methods for decision problems, it has also largely left the problem of developing stable DL classifiers open. 

\subsection{AI for inverse problems in imaging}

In much of the same way, the rapid developments in inverse problems in imaging have sometimes led to grand claims of robustness, new attack strategies, and conflicting results. There are, for example, no lack of works claiming to solve the issue of instabilities and/or hallucinations. To mention a few, the Nature publication \cite{Bo-18} promises \enquote{\textit{superior immunity to noise and a reduction in reconstruction artefacts compared with conventional hand-crafted reconstruction methods}}. However, this claim stands in stark contrast to the findings in \cite[Fig.\ 3]{Anetal-19}, showing that the proposed method is severely unstable to worst-case noise. Others have claimed that \enquote{\textit{the reconstructed images from NeRP are more robust and reliable, and can capture the small structural changes such as tumor or lesion progression.}}\cite{9788018}, or that   \enquote{\textit{The concept of deep image prior (DIP) [$\ldots$] is not affected by the aforementioned instabilities and hallucinations}}\cite{mean_field_pmlr21}, or \enquote{\textit{It is important to emphasize that the proposed GANCS scheme controls/avoids hallucination by modifying the conventional GAN in the following ways [$\ldots$]}}\cite{8417964}. 

At the same time, there have been numerous works demonstrating how easy it is to fool  modern reconstruction methods \cite{alaifari2022localized, Anetal-19,  Heckel21, genzel2020solving,  huang18,morshuis2022adversarial}, either by adding worst-case noise or adding tiny unseen details to the test data. Image reconstruction competitions such as the fastMRI challenges have also found that many of the best-performing algorithms can hallucinate \cite{fastmri19, fastmri20}, and that they produced overly smoothed images when higher noise levels were used \cite{johnson2021evaluation}. To us, this suggests that the question of how to avoid hallucinations is very much an open problem. 

\subsection{Towards robust AI}

For inverse problems, the conclusions of these studies and this paper are decidedly mixed: current approaches to training cannot ensure robust methods, and even if they do, the resulting methods may not offer state-of-the-art performance.  Furthermore, these are not rare events, able to be dismissed by all but a small group of theoreticians (recall Fig.\ \ref{fig:deep_mri_pert12}).  Should one therefore give up on the AI-based approaches for inverse problems in imaging?  Of course not. The potential for significant performance gains is too tempting to be dismissed. Consequently, there is at present a concerted effort to enhance robustness of AI-based methods.

The theoretical results presented in this paper shed some light on which approaches may be more profitable than others. One approach used in past work involves enforcing \textit{consistency}, i.e., $A \Psi(y) = y$. Another involves training with multiple sampling patterns, rather than a single sampling pattern. Unfortunately, neither approach helps prevent the mechanisms described our main theorems. A third approach is regularization and adversarial training (including with Generative Adversarial Networks). While this approach may mitigate against some of the mechanisms of our main theorems, they in turn introduce another instability involving setting the regularization parameter. We refer to \cite[\S 20.5]{adcock2021compressive} and references therein for a more detailed discussion of these approaches and their issues.

On a more positive note, our theoretical results emphasize that efforts that encourage learning stable reconstruction maps can mitigate both instabilities and hallucinations. As observed, simple strategies such as adding random noise to the measurements can encourage learning stable reconstruction maps. However, as discussed (see Fig.\ \ref{fig:3}), optimally balancing \textit{accuracy-stability} trade-off with this approach is by no means straightforward.

The fact that model-based methods are less susceptible to the effects predicted by our theoretical results suggest that hybrid approaches may be a good choice. Ideas from model-based methods are already frequently used in to design NN architectures via unrolling optimization algorithms (see, e.g., \cite{monga2021algorithm}, \cite[Chpt.\ 21]{adcock2021compressive}). However, this on its own is insufficient to avoid instabilities and hallucinations. Theorem \ref{thm:AIhal1} suggests that unrolling schemes where the architecture depends on the input may be worth pursuing, rather than a static architecture. Again, how best to balance the (generally) better robustness of model-based methods with the potential performance gains from data-driven methods remains very much an open problem.

Outside of these, some other approaches include using \textit{interval NNs (INNs)} \cite{intervalNN} or Bayesian approaches  \cite{barbano2021quantifying} to quantify the uncertainty in reconstructed images. While promising, it is currently unclear whether any of these approaches adequately resolve current concerns.

\subsection{Final note}
As we observed in \S \ref{ss:growing-concerns}, there are increasing warnings that, if issues such as hallucinations and instabilities cannot be brought under control, the eventual adoption of such techniques in safety-critical applications such as medical imaging may be less than the current optimism suggests.
Therefore, our hope is that the findings in this paper -- in particular, the crucial role that the nature of the forward operator plays in the generating hallucinations and instabilities -- will spur new research into devising better ways to design and train robust and reliable AI-based methods for inverse problems in imaging.

\appendix
\section{Additional information on the figures and experiments}\label{s:figures-experiments}
In this section we provide additional information on how the figures were created. In some of the figures we have trained the NNs ourselves. For these NNs, all accompanying code can be found on the GitHub page:
    \url{https://github.com/vegarant/troublesome_kernel}.
For the NNs which have been trained and published by others, we provide details on which data we have used.

\subsection{Fig.\ \ref{fig:1}}
Fig.\ \ref{fig:1} consists of four rows with different NNs tested on different data. In the first row, we downloaded the baseline model used in the 2020 fastMRI challenge \cite{fastmri20}. This baseline model is based on a variational NN\cite{sriram2020end}, and trained on the brain image dataset used in the competition. For further details on the training procedure and data used to train this model, we refer to \cite{fastmri20}. In our experiment, we used the pseudo-equispaced sampling pattern with 8X acceleration and the 12th slice from the \enquote{file\_brain\_AXT1PRE\_200\_6002079.h5} file in the validation dataset.

In the second row of Fig.\ \ref{fig:1}, we used a set of images published in the paper \cite{fastmri20} associated with the 2020 fastMRI challenge. These images are reconstructions created by the XPDNet \cite{ramzi2021xpdnet} trained for 4X acceleration.

In row three we consider the DFGAN-SISR model from \cite{qiao2021evaluation}. This NN has been trained on a dataset consisting of pairs of low and high-resolution microscopy images provided by \cite{qiao2021evaluation}. In our experiment, we used the DFGAN-SISR model trained for 2X upscaling of MicroTubules (MTs) images from a diffraction-limited wide-field view. For each specimen and each imaging modality in \cite{qiao2021evaluation}, the authors collected 50 pairs of low-resolution ($512 \times 512$ pixels) and high-resolution ($1024 \times 1024$ pixels) images. These images were then augmented using random cropping, horizontal/vertical flipping and rotation to generate a dataset of more than 20,000 image pairs of size $128\times 128$ and $256\times 256$. We have uploaded the relevant input and output images used in row 3 in Fig.\ \ref{fig:1} to the GitHub repository. For further details on the NN and training procedure, we refer to \cite{qiao2021evaluation}.

In the fourth row, the sampling operator $A \in \R^{m\times N}$ is a subsampled two-dimensional Hadamard transform with $m/N = 5\%$. The sampling pattern is shown in Fig.\ \ref{fig:pattern}. We trained a Tiramisu NN from \cite{genzel2020solving} with a learnable inverse layer initialized to the adjoint of the sampling operator before training. The NN was first trained for 15 epochs with random Gaussaian noise added to the measurements, then for another 10 epochs with noiseless measurements. For training, we used the Adam optimizer and a mean-squared-error loss function. The training data consisted of 10,000 images (of size $512 \times 512$ pixels) from the Endoplasmic Reticulum (ER) images acquired in \cite{qiao2021evaluation}. These 10,000 images were created by running the data-augmentation script provided in \cite{qiao2021evaluation} on the high-resolution images in this dataset. 

\subsection{Fig.\ \ref{fig:2}}
In Fig.\ \ref{fig:2}, we trained a NN $\Psi\colon \C^m\to \C^N$  to reconstruct the image $x = x_{\mathrm{br}} + x_{\mathrm{th}} + x_{\mathrm{mi}} \in \C^N$, along with 1200 other images from the fastMRI brain dataset. Here $x_{\mathrm{br}}$ is the brain image seen in the figure, $x_{\mathrm{th}}$ is the thumb detail and $x_{\mathrm{mi}}$ is the Mickey Mouse detail. The sampling operator $A \in \C^{m\times N}$  used in the experiment was a subsampled two-dimensional Fourier transform with $m/N=20\%$. The sampling pattern is shown in Fig.\ \ref{fig:pattern}.  The Mickey Mouse detail $x_{\mathrm{mi}}$ was designed so that $x_{\mathrm{mi}} \in \mathcal{N}(A)$. 

The ground truth images in the fastMRI dataset consist of magnitude images of size $320\times 320$. In our experiments, we resized these images to $256\times 256$ pixels and stored them as real-valued images with pixel values in the interval $[0,1]$. The detail $x_{\mathrm{mi}}$ is complex-valued, so any image with this detail is necessarily stored as a complex-valued image. To create the input data for the NN, we synthetically sampled these images with the sampling operator described above.

The NN $\Psi(y)= \varphi(A^*y)$ used a U-net $\varphi \colon \C^N\to \C^N$ combined with the adjoint $A^*$ of the sampling operator. The NN was trained in two phases. First, we trained the NN for 500 epochs with random Gaussian noise added to the measurements. Then we ran a fine-tuning phase for 100 epochs with noiseless measurements. We used the Adam optimizer with a gradually-decaying learning rate (see the GitHub page for details) and a mean-squared-error loss function. 

The image $x = x_{\mathrm{br}} + x_{\mathrm{th}} + x_{\mathrm{mi}}$ was part of the training set, whereas the images $x_{\mathrm{br}} + x_{\mathrm{th}}$ and $x_{\mathrm{br}}$ were not.  In Fig.\ \ref{fig:2} we have cropped the images to $180\times 180$ pixels to remove some dark areas surrounding the brain.

\subsection{Fig.\ \ref{fig:deep_mri_pert12}}
This figure consists of two experiments with two different NNs. We cover both in turn.

\begin{figure}
	\begin{center}
            \setlength{\tabcolsep}{1pt} 
		\begin{footnotesize}
            \begin{tabular}{@{}c@{}}
                Sampling pattern used for the NNs trained in Figures \ref{fig:1}-\ref{fig:deep_mri_pert12} \\
            \begin{tabular}{@{}>{\centering}m{0.23\textwidth}>{\centering}m{0.23\textwidth}>{\centering}m{0.23\textwidth}>{\centering\arraybackslash}m{0.23\textwidth}@{}} 
                 Fig.\ \ref{fig:1} 
               & Fig.\ \ref{fig:2} 
               & Fig.\ \ref{fig:3} 
               & Fig.\ \ref{fig:deep_mri_pert12} \\
                 $N=512\times 512$, $m/N = 5\%$ 
               & $N=256\times 256$, $m/N = 20\%$ 
               & $N=256\times 256$, $m/N = 17\%$ 
               & $N=256\times 256$, $m/N = 33\%$ \\ 
				 \includegraphics[width=\linewidth]{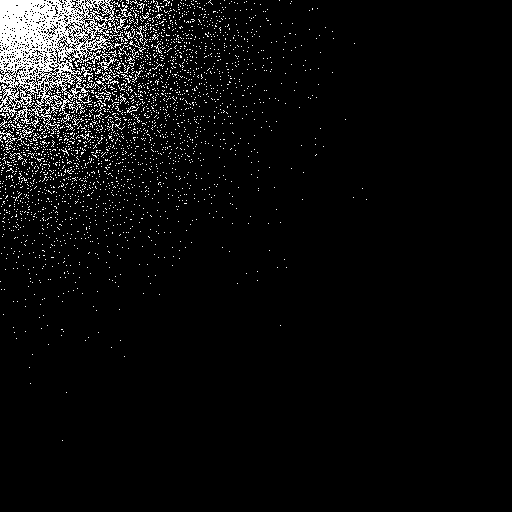}
                &\includegraphics[width=\linewidth]{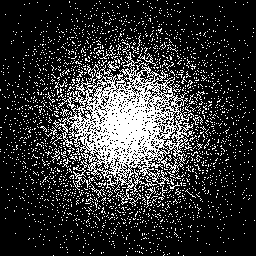}
                &\includegraphics[width=\linewidth]{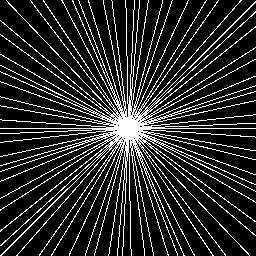}
                &\includegraphics[width=\linewidth]{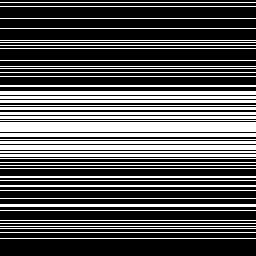} \\
            \end{tabular}
            \end{tabular}
        \end{footnotesize}
    \end{center}
    \caption{\label{fig:pattern} In Figures \ref{fig:1}-\ref{fig:deep_mri_pert12} we have trained our own NNs. In this figure we visualize the different sampling patterns used for training of these NNs. Note that the leftmost sampling pattern is used for Hadamard sampling, whereas the three others are used for Fourier sampling.}
\end{figure}

In the four leftmost images, we consider the DeepMRI-Net from \cite{SchCaH-17}. This NN is composed of a cascade of U-Nets and data consistency layers. It has been trained on cardiac images, such as the one shown in the figure. The sampling operator $A $ is a subsampled two-dimensional Fourier transform, whose sampling pattern is shown in Fig.\ \ref{fig:pattern}. The noise vector $v$ used in the experiment is defined as $v = A^* e$, where $e \in \C^m$ is a zero-mean complex-valued Gaussian noise vector. Since the mean of a Gaussian random variable is unchanged by a linear transformation, the noise vector $v$ has zero mean as well.

\begin{figure}
    \begin{center}
	\setlength{\tabcolsep}{5pt} 
        \begin{footnotesize}
        \begin{tabular}{@{}>{\centering}m{0.40\textwidth}>{\centering\arraybackslash}m{0.40\textwidth}@{}}
        Original image used in Fig.\ \ref{fig:deep_mri_pert12}
            & Image used for worst-case comp.\ Fig.\ \ref{fig:deep_mri_pert12} \\
         \includegraphics[width=1\linewidth]{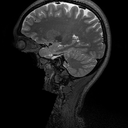} 
        &\includegraphics[width=1\linewidth]{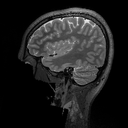} 
    \end{tabular}
    \end{footnotesize}
    \end{center}
    \caption{\label{fig:worst-case} Left: the original image $x$ in Fig.\ \ref{fig:deep_mri_pert12}. Right: the image that was used to compute the worst-case perturbations $e_{\mathrm{pert}, j}^1$ that were subsequently used as the mean of the Gaussain distributions in Fig.\ \ref{fig:deep_mri_pert12}. }
\end{figure}

The four rightmost images are from \cite{antun2023ai}. Code is available at \url{https://github.com/vegarant/am_AI_hallucinating}. Here we consider the AUTOMAP network from \cite{Bo-18}. This NN was trained by the authors of \cite{Bo-18} on brain images from the MGH--USC dataset \cite{Fan16}. It was trained using Fourier sampling with 60\% subsampling. In \cite{antun2023ai}, the perturbations 
 \[e_j = e^{1}_{\mathrm{pert},j} + e^{2}_{\mathrm{pert},j} \in \C^m, \quad \text{for } j = 0,1,2,3.\]
Here $e^{1}_{\mathrm{pert},0} = 0$ and $e^{2}_{\mathrm{pert},j}$, $j=0,\ldots, 3$ are zero-mean Gaussian vectors. The vectors $e^{1}_{\mathrm{pert},j}$, $j=1,2,3$, are worst-case noise vectors computed for an image that differs from the image $x$ used in the experiment. This makes the mean of the Gaussian noise vector $e_j$ image independent. The image used to compute the worst-case perturbations $e_{\mathrm{pert},j}^1$ is shown in Fig.\ \ref{fig:worst-case}.

\subsection{Fig.\ \ref{fig:new_fig}}
In Fig.\ \ref{fig:new_fig}, we trained a NN to reconstruct CT images from Radon measurements. For the sampling operator, we used MATLAB's implementation of the Radon transform, and sampled 50 equally spaced angles. The choice of considering 50 angles is inspired by the seminal work  \cite{JinMcUn-17}. Since we considered images with dimensions $256\times 256$, the sampling operator had dimensions $N=256^2 = 65,536$ and $m=18,350$.

In this experiment we used the 100 CT images from the cancer imaging archive \cite{clark2013cancer} that have been made available at \url{kaggle.com} (See:  \url{https://www.kaggle.com/datasets/kmader/siim-medical-images/}). Due to the small size of this dataset, we trained the NN in two stages. In the first stage, we pretrained the NN on the 25,000 ellipses images used to train the NNs in Fig.\ \ref{fig:3}. Then we fine-tuned the NN on 95 CT images from the dataset mentioned above. Among the 95 images used for training was the image $x+x_{\mathrm{Det}}$, where $x_{\mathrm{Det}}$ is the \enquote{SIAM} detail seen in the figure. The clean image $x$ was not part of the training data. The detail $x_{\mathrm{Det}}$ was designed so that $0 < \|Ax_{\mathrm{Det}}\|_{\ell^2} \ll \|Ax\|_{\ell^2}$. Specifically, we computed $x_{\mathrm{Det}}$ as $(I - (A^*A + \alpha I)^{-1}A^*A)\tilde x_{\mathrm{Det}}$, where $I$ is the identity matrix, $\alpha = 0.01$ and $\tilde x_{\mathrm{Det}}$ is a black image with the \enquote{SIAM} text feature.

The architecture of the trained NN can be written as $\Psi(y) = \varphi(Ky)$, where $\varphi\colon \R^{N}\to \R^N$ is a learnable U-Net and $K\in \R^{N\times m}$ is the matrix of a (non-learnable) Filtered Backprojection (FBP) with a \enquote{Ram-Lak}-filter. The \enquote{Ram-Lak}-filter was chosen as it is the default filter in MATLAB. We trained the network for 60 epochs in the pretaining phase and for 520 epochs in the fine-tuning phase. We used a larger number of epochs in the fine-tuning phase than in the pretraining phase, as the amount of training data was substantially smaller in the fine-tuning phase. During all training epochs we used a mean-squared-error loss function and noiseless measurements.  

\begin{figure}
	\begin{center}
	\begin{tabular}{@{}>{\centering \small }m{0.3\textwidth}>{\centering \small \arraybackslash}m{0.3\textwidth}@{}}
        $x + x_{\mathrm{Det}}$ & $x$  \\
        \includegraphics[width=\linewidth]{plots/sample_00005_gt.png} & 
        \includegraphics[width=\linewidth]{plots/sample_00101_gt.png}  \\ 
        $K (A(x + x_{\mathrm{Det}}))$ & $K (Ax)$ \\ 
        \includegraphics[width=\linewidth]{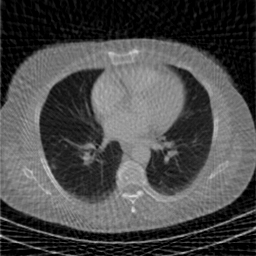} & 
                \includegraphics[width=\linewidth]{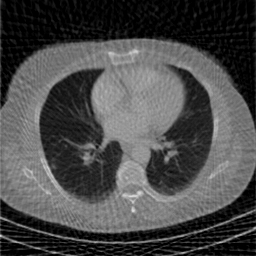} 
          \end{tabular}
    \end{center}
    \caption{\label{fig:not-FBP} The detail transfer shown in Fig.\ \ref{fig:new_fig} is not due to the FBP operator $K$. This figure shows the reconstruction of the images $x$ and $x + x_{\mathrm{Det}}$ by $K$. Since $x_{\mathrm{Det}}$ lies close to the kernel of $A$ and the FBP operator is only mildly ill-conditioned, the detail $x_{\mathrm{Det}}$ in the left image is imperceivable when reconstructed via $K$ alone. This implies that the hallucination of $x_{\mathrm{Det}}$ shown Fig.\ \ref{fig:new_fig} must be due to the NN $\varphi$.}
\end{figure}

By construction, the NN $\Psi$ acts by denoising the FBP image $K y$ using the U-Net $\varphi$. It is important to note that the detail transfer shown in Fig.\ \ref{fig:new_fig} stems from the action of the NN $\varphi$, not the FBP operator $K$. We illustrate this in Fig.\ \ref{fig:not-FBP}.

\subsection{Fig.\ \ref{fig:3}}
In Fig.\ \ref{fig:3} we trained three NNs on a dataset consisting of 25,000 ellipses images of size $256\times 256$. All the NNs have the same architecture, consisting of a cascade of U-Nets and learnable data-consistency layers. This architecture is identical to the ItNet in \cite{genzel2020solving} and resembles the state-of-the-art architecture used in \cite{pmlr-v162-genzel22a} for CT reconstruction. 
In our experiments we used a subsampled Fourier transform as our sampling operator $A \in \C^{m\times N}$, with $m/N \approx 17\%$ and the sampling pattern shown in Fig.\ \ref{fig:pattern}, consisting of 40 radial lines. 

We trained the NNs using a mean-squared-error loss function and noisy inputs. 
The technique of adding noise to the input is known as jittering in the machine learning literature. We refer to the intensity of the noise as the jittering level. At jittering level $p > 0$ we drew -- for each noiseless measurement vector $y$ and for each epoch of training -- a number $s \sim \mathrm{Unif}([0,p])$ from the uniform  distribution on $[0,p]$ and a noise vector $e\sim \mathbb{C}\mathcal{N}(0, \tfrac{1}{2}I)$, which we then used to generate the noisy measurements 
\[
\tilde{y} \gets y + \frac{s}{\sqrt{m}}e. 
\]
The NNs were trained in different stages and for varying levels of jittering. In the first stage, we trained a simple U-Net to reconstruct the images using a jittering level of $p=10$. In the second stage, we trained NNs with the architecture described above consisting of a cascade of U-Nets and data-consistency layers. The NN weights of these U-Nets were initialized using the U-Net from the first stage. At this stage, two of the NNs were trained at jittering level $10$, whereas the last NN was trained without any jittering. In Fig.\ \ref{fig:3}, we refer to the NN trained without any jittering as the NN trained without noise. Furthermore, one of the NNs trained at jittering level $10$ is referred to as the NN trained with high levels of noise. The weights of the other NN trained at jittering level $10$ were used as a warm start for the NN trained with low levels of noise. This NN was trained another time, in a third stage, using a jittering level of $1/10$. The training strategy of these three networks is identical to the training procedure used in \cite[Fig.\ 14]{genzel2020solving}.

To test the accuracy of the trained NNs, we insert the \enquote{SIAM} text feature in one of the test images and reconstruct this image from noiseless measurements. None of the NNs have been trained on images with text in them, yet two of the NNs reconstruct this text feature with high accuracy. The NN that fails to reconstruct this text feature is the one trained at the highest noise level. 

In the graph in Fig.\ \ref{fig:3} we test the stability of the NNs towards worst-case noise at different noise levels. This is done as follows. For each NN, we consider 10 images and the noise levels $\eta = 0, 0.02, 0.04, 0.06, 0.08$. For each image $x$, noise level $\eta$ and NN $\Psi$, we compute a worst-case perturbation
\[ e^{\mathrm{worst}} \in \mathrm{argmax}_{e\in \C^m} \|\Psi(Ax + e) - x\|_{\ell^2}^{2} \quad\text{subject to}\quad \|e\|_{\ell^2} \leq \eta\|x\|_{\ell^2} \]
using a projected gradient descent algorithm. The figure shows the average relative reconstruction error at each noise level. The wider shaded areas of each colour indicate the boundaries for one standard deviation of the relative reconstruction error for the considered dataset of 10 images.

\section{Model-based reconstructions, hallucinations and instabilities}\label{s:model-based-CS}

In this section, we show that the hallucinations and instabilities implied by Theorems \ref{thm:AIhal2}, \ref{thm:AIhal1} and \ref{thm:locallipcte} generally cannot arise in the case of model-based reconstructions based on $\ell^1$-minimization, subject to suitable conditions on the matrix $A$.

\subsection{Sparse recovery via $\ell^1$-minimization}

Specifically, let $H \in \bbC^{N \times N}$ be a unitary \textit{sparsifying} transform (e.g., a discrete wavelet transform), $1 \leq s \leq N$, $\delta > 0$ and consider the model class
\bes{
\cM_1 = \left \{ x \in \bbC^N : \sigma_s(H x)_{\ell^1} / \sqrt{s} \leq \delta \right \}
} 
consisting of all approximately $s$-sparse vectors.
Here 
$$
\sigma_s(c)_{\ell^1} = \min \{ \nm{c - z}_{\ell^1}: \text{$z$ is $s$-sparse} \}
$$
is the $\ell^1$-norm best $s$-term approximation error of a vector $c \in \bbC^N$. Recall that a vector is \textit{$s$-sparse} if it has at most $s$ nonzero entries. The use of $\ell^1$-norm in this definition is so we may apply compressed sensing theory below. We note also that the condition that $H$ is unitary may be relaxed. With some further effort, one can also consider redundant sparsifying transforms such as the discrete gradient operator (see, e.g., \cite[Chpt.\ 17]{adcock2021compressive} and \cite{neyra-nesterenko2023stable}).

Given this choice of $\cM_1$, the next step in a model-based reconstruction is design a reconstruction map $\Psi$ that performs well over $\cM_1$. It is well-known that this can be done by solving a suitable $\ell^1$-minimization problem. Specifically, given $\eta \geq 0$ we now define the reconstruction map $\Psi : \bbC^m \rightarrow \bbC^N$, $y \mapsto \hat{x}$, where $\hat{x}$ is any minimizer of the $\ell^1$-minimization problem
\be{
\label{QCBP}
\min_{z \in \bbC^N} \nm{H z}_{\ell^1}\ \text{subject to }\nm{A z - y}_{\ell^2} \leq \eta.
}
The problem \eqref{QCBP} is known as \textit{Quadratically Constrained Basis Pursuit (QCBP)}.
Note that one may also consider an unconstrained, LASSO-type optimization problem with no additional difficulties.  Observe that $\Psi$ is not, technically speaking, a well-defined map, since \eqref{QCBP} in general has infinitely many minimizers. This can be treated by either considering a multivalued map, or by fixing a specific minimizer of \eqref{QCBP} (e.g., the one with minimal $\ell^2$-norm). Practically, one could consider $\Psi$ as the output of some algorithm that solves \eqref{QCBP} to some tolerance. It is, however, of little consequence to what follows, since the various error bounds will hold for all minimizers of \eqref{QCBP}.

\subsection{Accurate and stable recovery via the rNSP}

We now describe conditions on $A$ that ensure good recovery when solving \eqref{QCBP}. Specifically, we suppose that $A H^* \in \bbC^{m \times N}$ satisfies the \textit{robust Null Space Property (rNSP)} of order $1 \leq s \leq N$ with constants $0 < \rho < 1$ and $\gamma > 0$, i.e.,
\bes{
\nm{c_{S}}_{\ell^2} \leq \frac{\rho}{\sqrt{s}} \nm{c_{S^c}}_{\ell^1} + \gamma \nm{A H^* c}_{\ell^2}
}
for all $S \subseteq \{1,\ldots,N\}$, $|S| \leq s$ and $c \in \bbC^N$ (see, e.g., \cite[Defn.\ 5.14]{adcock2021compressive}). Here $c_{S}$ is the vector with $i$th entry equal to the $i$th entry of $c$ for $i \in S$ and zero otherwise. It is well-known that this property guarantees accurate and stable recovery of approximately sparse vectors via \eqref{QCBP}. For instance, given $x \in \bbC^N$ and $y = A x + e \in \bbC^m$, then 
\be{
\label{CS-error-bound}
\nm{\hat{x} - x}_{\ell^2} \leq C_1 \frac{\sigma_s(H x)_{\ell^1}}{\sqrt{s}} + C_2 \eta,
}
for any minimizer $\hat{x}$ of \eqref{QCBP} with $\eta \geq \nm{e}_{\ell^2}$. Here $C_1$ and $C_2$ are constants depending on $\rho$ and $\gamma$ only. See, e.g., \cite[Thm.\ 5.17]{adcock2021compressive}.

\subsection{The case of Theorems \ref{thm:AIhal2} and \ref{thm:AIhal1}}

We are now ready to consider the implications our main results for this model-based reconstruction map. First, consider the setup of Theorem \ref{thm:AIhal2}. In general, we expect \eqref{eq:NN_exists} to hold only when $x + x_{\mathrm{Det}} \in \cM_1$, since, due to \eqref{CS-error-bound}, $\cM_1$ is the class over which this model-based map is guaranteed to perform well. Then Theorem \ref{thm:AIhal2} implies that
\be{
\label{x-det-rec-CS}
\Psi(A x ) \approx x + x_{\mathrm{Det}}.
}
We now consider two cases. Suppose first that $x \in \cM_1$. In this case, $\Psi$ must also recover $x$ well, i.e., $\Psi(A x) \approx x$. Therefore, we must have that $x_{\mathrm{Det}}$ is small. In other words, the reconstruction map does not hallucinate. Conversely, if $x \notin \cM_1$ then in general we will not recover $x$ well via $\Psi$. Hence the fact that \eqref{x-det-rec-CS} occurs is not indicative of a hallucination, since we have no expectation of being able to recover $x$ well in the first place.
In general, a hallucination only exists if one's expectation is to recover the detail-free image well in the first place.

In summary, as long as $A$ satisfies suitable conditions (e.g., the rNSP), model-based reconstruction maps involving sparsity and $\ell^1$-minimization are unlikely to hallucinate. Note that similar considerations apply in the case of Theorem \ref{thm:AIhal1}. Arguing in a more informal sense, model-based maps also tend to fail in more recognizable ways. As suggested by \eqref{CS-error-bound}, the error of such a map generally behaves like the $s$-term error in the given transform. So if one uses a wavelet basis, for instance, the error will present as typical wavelet artefacts, which are generally nonphysical in nature. In other words, it is very unlikely for the map to hallucinate highly structured artefacts like the `Mickey Mouse' or `Thumb' details shown in Fig.\ \ref{fig:2}.

\subsection{The case of Theorem \ref{thm:locallipcte}}

Now consider Theorem \ref{thm:locallipcte} and let $\nm{\cdot}$ and $\vertiii{\cdot}$ both equal the Euclidean norm $\nm{\cdot}_{\ell^2}$ for convenience. Suppose that \eqref{eq:cond_instability} and \eqref{eq:cond_instability2} hold. We also assume that $x \in \cM_1$ for some $\delta \leq \eta$, since this is generally the setting in which one expects $\nm{\Psi(A x) - x}_{\ell^2}$ to be small. Let $e = A x' - A x$ so that $A x' = A x + e$. Then, by the triangle inequality,
\bes{
\nm{x - x''}_{\ell^2} \leq \nm{x - \Psi(A x') }_{\ell^2} + \nm{\Psi(A x') - x''}_{\ell^2}.
}
We now apply \eqref{CS-error-bound}, \eqref{eq:cond_instability2} and \eqref{eq:cond_instability} to get
\bes{
\nm{x - x''}_{\ell^2} \leq C_1 \delta  + C_2 \eta + \eta \leq (C_1+C_2+1)\eta.
}
Therefore $x$ and $x''$ must be close for small $\eta$. This voids the main conclusions of the theorem. In other words, the compressed sensing-based reconstruction map will generally not be unstable or hallucinate in the ways described in this theorem, as long as $A$ satisfies appropriate conditions (i.e., the rNSP).

The key point is that $\Psi$ and $A$ have been designed jointly in terms of $\cM_1$ to ensure good recovery of vectors in $\cM_1$ well from noisy measurements. Thus, one cannot have wildly different reconstructions of $x$ and $x'$ whenever $x \in \cM_1$ and $A x' \approx A x$. By contrast, a different type of reconstruction map can easily be trained to perform in this way, thereby leading to instabilities.

{\footnotesize
\bibliographystyle{abbrv}
\bibliography{references}
}

\end{document}